\RequirePackage{fix-cm}
\documentclass[twocolumn]{svjour3}          
\smartqed  

\usepackage{placeins}
\usepackage{tabularx}
\usepackage{cite}
\usepackage[pdftex]{graphicx}
\usepackage{ragged2e}
\usepackage[tight,footnotesize]{subfigure}
\usepackage{rotating}
\usepackage{graphicx}
\usepackage{amsmath,amssymb} 
\usepackage{array}
\usepackage{multirow}
\usepackage{colortbl}
\usepackage{amsfonts}
\usepackage{pifont}
\usepackage{xspace}
\usepackage{etoolbox}
\usepackage{overpic}
\usepackage{color}
\usepackage{microtype}
\usepackage{bm}
 \usepackage{booktabs}
\usepackage{caption}
\usepackage{xcolor}
\definecolor{DarkGold}{HTML}{B8860B}
\definecolor{GingerYellow}{HTML}{D5B15F} 
\definecolor{DeepAmber}{HTML}{D97706}

\newcommand{\orcid}[1]{\href{https://orcid.org/#1}{\includegraphics[width=10pt]{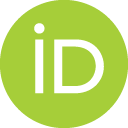}}}

\def\etal{{\em et al}}

\usepackage{hyperref}
\hypersetup{breaklinks=true,citecolor=blue, colorlinks}

\graphicspath{{./Imgs/}}
\DeclareGraphicsExtensions{.pdf,.jpg,.png}

\usepackage{silence}
\journalname{Research Article}

\begin{document}

\title{YUV20K: A Complexity-Driven Benchmark and Trajectory-Aware Alignment Model for Video Camouflaged Object Detection}

\titlerunning{YUV20K}        

\author{ Yiyu Liu~$^\dag$\and
   Shuo Ye~$^\dag$ \and 
   Chao Hao \and
   Zitong Yu~$^{*}$\orcid{https://orcid.org/my-orcid?orcid=0000-0003-0422-6616} \and \\
}

\authorrunning{Yiyu Liu \etal} 

\institute{
$^\dag$ The authors contributed equally to this work.\\
$*$ Correspondence: yuzitong@gbu.edu.cn\\
Yiyu Liu, Shuo Ye, Chao Hao and Zitong Yu are with the Department of Computer Science, Great Bay University, Dongguan, Guangdong, 523000, China (Email: yiyuliu@gbu.edu.cn, shuoye.ke@gmail.com, chaohao@gbu.edu.cn and yuzitong@gbu.edu.cn). 
}

\date{Received: date / Accepted: date}

\maketitle

\begin{abstract}
Video Camouflaged Object Detection (VCOD) is currently constrained by the scarcity of challenging benchmarks and the limited robustness of models against erratic motion dynamics. Existing methods often struggle with Motion-Induced Appearance Instability and Temporal Feature Misalignment caused by complex motion scenarios. To address the data bottleneck, we present YUV20K, a pixel-level annoated complexity-driven VCOD benchmark. Comprising 24,295 annotated frames across 91 scenes and 47 kinds of species, it specifically targets challenging scenarios like large-displacement motion, camera motion and other 4 types scenarios. On the methodological front, we propose a novel framework featuring two key modules: Motion Feature Stabilization (MFS) and Trajectory-Aware Alignment (TAA). The MFS module utilizes frame-agnostic Semantic Basis Primitives to stablize features, while the TAA module leverages trajectory-guided deformable sampling to ensure precise temporal alignment. Extensive experiments demonstrate that our method significantly outperforms state-of-the-art competitors on existing datasets and establishes a new baseline on the challenging YUV20K. Notably, our framework exhibits superior cross-domain generalization and robustness when confronting complex spatiotemporal scenarios. Our code and dataset will be available at \url{https://github.com/K1NSA/YUV20K}


\end{abstract}

\begin{figure*}[ht!p]
    \centering
    
    \includegraphics[width=\linewidth]{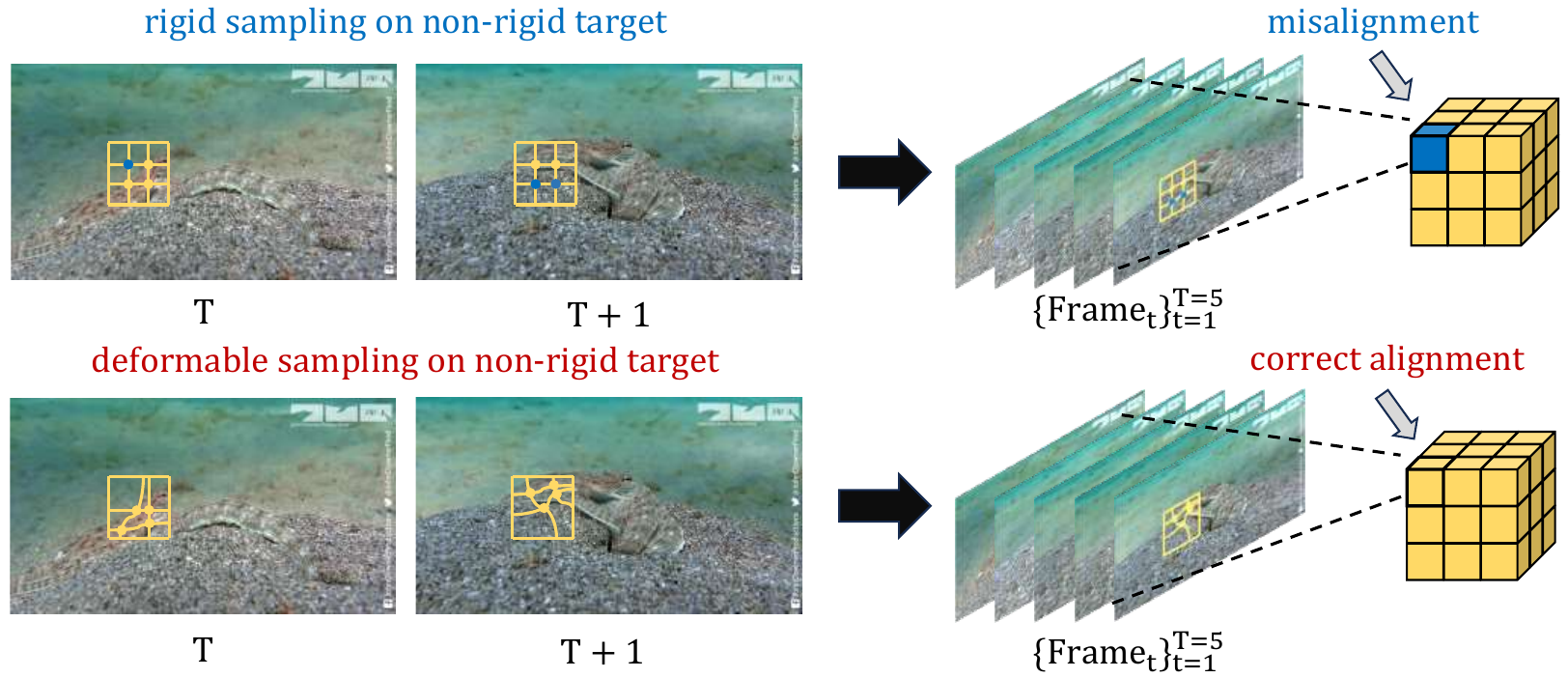}
    
    \caption{\textbf{Top: Misalignment of rigid sampling convolutions.} When handling non-rigid moving objects, standard sampling convolutions with fixed sampling grids fail to adapt to the target's continuous deformation. This mismatch causes the network to inevitably sample irrelevant background regions, indicated by the \textcolor{blue}{blue} points, rather than the target features highlighted in \textcolor{DarkGold}{yellow}. This leads to severe background noise accumulation over time. \textbf{Bottom: Alignment via deformable sampling.} By dynamically adjusting the sampling grid, our deformable strategy faithfully tracks the target's features across frames, ensuring all sampled points remain \textcolor{DarkGold}{yellow}. This actively avoids background interference, mitigating noise accumulation and ensuring accurate spatiotemporal feature alignment.}
    \label{fig:motivation}
\end{figure*}

\section{Introduction}
Camouflaged Object Detection (COD) aims to identify and segment objects that exhibit high visual similarity to their surroundings\cite{fan2021concealed}. Unlike generic object detection that focuses on salient targets, COD tackles scenarios where the foreground and background share homogeneous textures, luminance, and patterns. Consequently, this task requires models to effectively decouple these confusing visual cues and accurately localize subtle structural boundaries. Consequently, COD not only facilitates practical applications in medical image segmentation\cite{fan2020pranet,fan2020inf}, agricultural pest monitoring\cite{rustia2020application} but also drives the development of more robust and perceptive computer vision systems.

Image COD (ICOD) focuses on detecting camouflaged targets within static images. Since these targets exhibit extreme spatial consistency with their surroundings, ICOD models are restricted to relying solely on static appearance cues, making the detection highly challenging.
Consequently, motion cues for video camouflaged object detection effectively break the target-background homogeneity\cite{lamdouar2020betrayed, yang2021self}, transforming the ill-posed static problem into a tractable spatiotemporal inference task.

However, while temporal motion cues are crucial for breaking camouflage, they simultaneously introduce a new layer of complexity that has been overlooked. Existing methods typically operate under the assumption of stable motion, characterized by linear trajectories and rigid poses\cite{lamdouar2020betrayed}. In contrast, real-world camouflaged creatures exhibit highly complex motion dynamics, such as an octopus squeezing through a narrow crevice (non-rigid deformation) or a predator launching a lethal strike (large-displacement motion). Overlooking these scenarios introduces two fundamental impediments to effective segmentation: Motion-Induced Appearance Instability and Temporal Feature Misalignment. First, in the spatial domain, complex motion induces appearance instability. Rapid motion results in blur that obliterates texture details\cite{perazzi2016benchmarkmotionblur}, while slow motion renders motion cues imperceptible, hindering effective figure-ground separation\cite{yang2021uncertainty}. Second, in the temporal domain, irregular motion leads to severe feature misalignment. Camouflaged targets frequently undergo extreme non-rigid deformations or erratic displacements, which fundamentally violate the rigid grid assumptions of standard aggregation modules (e.g., 3D convolutions)\cite{wang2019edvr}. As visually demonstrated in Fig.~\ref{fig:motivation} (Top), when handling such non-rigid targets, standard rigid sampling convolutions fail to adapt to continuous deformations. Consequently, the fixed sampling grid inevitably captures irrelevant background regions, causing severe noise accumulation and spatial misalignment across frames.

To address Motion-Induced Feature Instability, we propose the Motion Feature Stabilization module (MFS). Within this module, we introduce a set of Semantic Basis Primitives (SBPs) initialized via a Gaussian distribution. Designed as frame-agnostic learnable parameters, these primitives remain invariant across dynamic input frames. By injecting them into the feature stream, we augment the semantic representation, expand the feature space. Functioning as robust anchors during global query operations, the SBPs enable the model to effectively lock onto the camouflaged target within the global feature space. 
To resolve Temporal Feature Misalignment, driven by the dynamic alignment depicted in Fig.~\ref{fig:motivation} (Bottom), we propose the Trajectory-Aware Alignment (TAA) module. Drawing inspiration from deformable paradigms\cite{dcnzhu2020deformabledetr,wang2023internimage}, TAA moves beyond rigid aggregation. Specifically, it leverages temporal features to explicitly predict pixel-wise motion offsets, effectively modeling the target's trajectory. These offsets are then applied to warp the spatial sampling grid, ensuring that features are aggregated strictly along the object's actual deformation path. Unlike standard sampling which performs indiscriminate averaging on a fixed grid, our trajectory-guided deformable sampling precisely aligns distinct features, significantly optimizing spatiotemporal feature integrity.

\begin{table}[t]
    \centering
    \caption{Comparison of YUV20K with existing VCOD datasets. \textbf{YUV20K} provides the most comprehensive annotations.}
    \label{tab:dataset_comparison}
    \resizebox{\linewidth}{!}{
        \begin{tabular}{l | c c c | c c c } 
            \toprule
            \textbf{Dataset} & \textbf{Venue} & \textbf{videos/frames} & \textbf{species} & \textbf{Class} & \textbf{B.Box} & \textbf{Mask} \\
            \midrule
            CAD\cite{bideau2016scad} & ECCV'16 & 9/839 & 6 & \checkmark & $\times$ & \checkmark \\
            
            MOCA\cite{lamdouar2020betrayed} & ACCV'20 & 141/37,250 & 67 & $\times$ & \checkmark & $\times$ \\ 
            
            MOCA-Mask\cite{cheng2022implicit} & CVPR'22 & 87/22,939 & 44 & \checkmark & $\times$ & \checkmark 
            \\

            \midrule
            \textbf{YUV20K} & - & \textbf{91/24,295} & \textbf{47} & \checkmark & \checkmark & \checkmark \\
            \bottomrule
        \end{tabular}
    }
\end{table}

Beyond algorithmic design, the field is constrained by the scarcity of benchmarks that fully capture the complexity of VCOD\cite{xiao2024survey}. While pioneering datasets have established a solid baseline, capturing the full spectrum of erratic dynamics inherent to wild biological behaviors remains an open challenge. This leaves a critical gap in representing challenging scenarios, such as large-displacement dashes, severe non-rigid deformations and multi-object interactionss gap, we present YUV20K. Comprising 91 clips, YUV20K is positioned as a challenging and scenario diverse benchmark.  By including a higher proportion of complex scenarios (e.g.,  large displacement motion, camera motion), YUV20K aims to expose the limitations of current models and drive the development of more robust spatiotemporal reasoning systems.

Our main contributions are summarized as follows:
\begin{itemize}
    \item We propose the Motion Feature Stabilization (MFS) module to address motion-induced appearance instability. By introducing frame-agnostic Semantic Basis Primitives as anchors, MFS effectively stabilizes semantic features against blur and rapid changes, enabling robust target locking even under erratic motion dynamics.

    \item We design the Trajectory-Aware Alignment (TAA) module to resolve temporal feature misalignment. Moving beyond rigid grid assumptions, TAA utilizes trajectory-guided deformable sampling to predict pixel-wise motion offsets, ensuring precise feature aggregation along the object's complex deformation path.
    \item We present YUV20K, a complexity-driven benchmark comprising 91 clips that captures the full spectrum of wild biological behaviors. Extensive experiments demonstrate that our method achieves state-of-the-art performance on this challenging dataset and existing benchmarks, setting a new baseline for robust video camouflaged object detection.
\end{itemize}

\section{Related Work}

\subsection{Camouflaged Object Detection}

Image-based COD aims to identify concealed objects from a single static RGB image. This task is inherently challenging because camouflaged targets share high intrinsic similarities with their backgrounds. Biological studies\cite{hall2013camouflage} reveal that predatory animals typically employ a two-step strategy to break camouflage: first scanning the environment to locate potential prey, followed by precise identification. Inspired by this mechanism, Fan et al.\cite{fan2020camouflaged,fan2021concealed} proposed a coarse-to-fine framework that initially generates a localization map to discover the target, which is subsequently refined into a precise pixel-level segmentation mask.

Another line of research attempts to tackle the high visual similarity by exploring alternative feature spaces and relational modeling. For instance, recent works \cite{cong2023frequency, sun2024frequency} utilize frequency cues, fusing complementary information from the frequency domain with spatial representations to discern subtle differences. Meanwhile, methods like MGL\cite{zhai2021mutual} introduce mutual graph learning to explicitly decouple the underlying topological relationship between the camouflaged object and its surroundings. With the development of deep learning architectures, recent methods rely heavily on multi-scale feature integration. Models like ZoomNet\cite{pang2022zoom} and FSPNet\cite{huang2023feature} effectively model context at both global and local scales, capturing scale-specific semantics. Despite these advancements, image-based methods rely solely on static appearance cues, making them vulnerable to complex scenes where boundaries remain completely ambiguous without temporal information.

\subsection{Video Camouflaged Object Detection}

Compared to static images, motion cues between video frames provide critical information for breaking camouflage. Existing VCOD strategies primarily focus on how to effectively utilize spatial-temporal information, which can be broadly categorized into explicit and implicit motion modeling.

Explicit motion methods typically employ off-the-shelf optical flow estimators (e.g., RAFT\cite{teed2020raft}) to separate moving foregrounds from backgrounds. Lamdouar et al.\cite{lamdouar2020betrayed} take optical flow and difference images as inputs, designing a differentiable registration module for background alignment and a motion segmentation module to discover moving objects. Similarly, recent two-stream architectures \cite{zhang2025explicitemip} simultaneously conduct camouflaged object segmentation and explicit optical flow estimation. However, explicitly extracting optical flow introduces several challenges. It struggles when the camouflaged object remains stationary or under severe camera motion. Furthermore, due to the repetitive textures of camouflaged objects, pixel-level explicit motion estimation is often noisy and incurs considerable computational expense\cite{hui2024endow}.

To overcome the limitations of explicit optical flow, recent works have shifted toward learning implicit inter-frame feature correspondences for temporal alignment. SLT-Net \cite{cheng2022implicit} introduced a model that implicitly captures motion over both short and long temporal intervals, ensuring temporal consistency while directly predicting pixel-level masks. To further reduce architectural redundancy, recent trends aim to unify static and dynamic processing. For example, ZoomNeXt\cite{pang2024zoomnext} uses a difference-aware routing mechanism to adaptively propagate inter-frame temporal cues. Additionally, models like SAM-PM\cite{meeran2024sampm} explore spatio-temporal cross-attention mechanisms to enforce temporal consistency across consecutive frames.

Despite these improvements, existing methods still struggle with motion-induced appearance instability and temporal feature misalignment under complex trajectories. Furthermore, while pioneering datasets like MoCA\cite{lamdouar2020betrayed}, CAD\cite{bideau2016scad}, and MoCA-Mask\cite{cheng2022implicit} have significantly advanced VCOD, they primarily serve as foundational benchmarks. As algorithms evolve, evaluating spatial-temporal reasoning under highly complex dynamic scenes (e.g., severe environmental degradation and sustained complex trajectories) demands benchmarks with larger scales and higher annotation densities.

\begin{figure}[ht!]
    \centering
    \includegraphics[width=\linewidth]{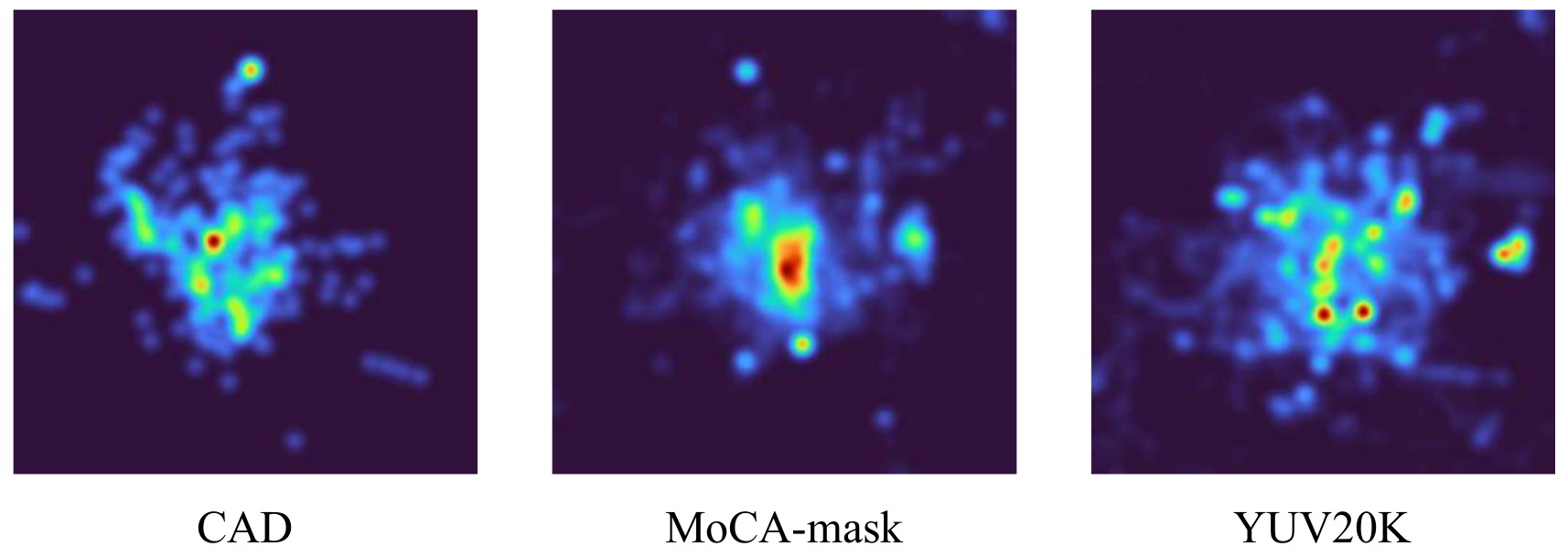}
    \caption{\textbf{Spatial distribution density maps.} Compared to CAD and MoCA-Mask, where targets tend to appear in the central area, our YUV20K provides a broader and more uniform spatial distribution. This diverse placement introduces greater challenges for robust segmentation.}
    \label{fig:density_comparison}
\end{figure}

\begin{figure*}[t!]
    \centering
    \includegraphics[width=\textwidth]{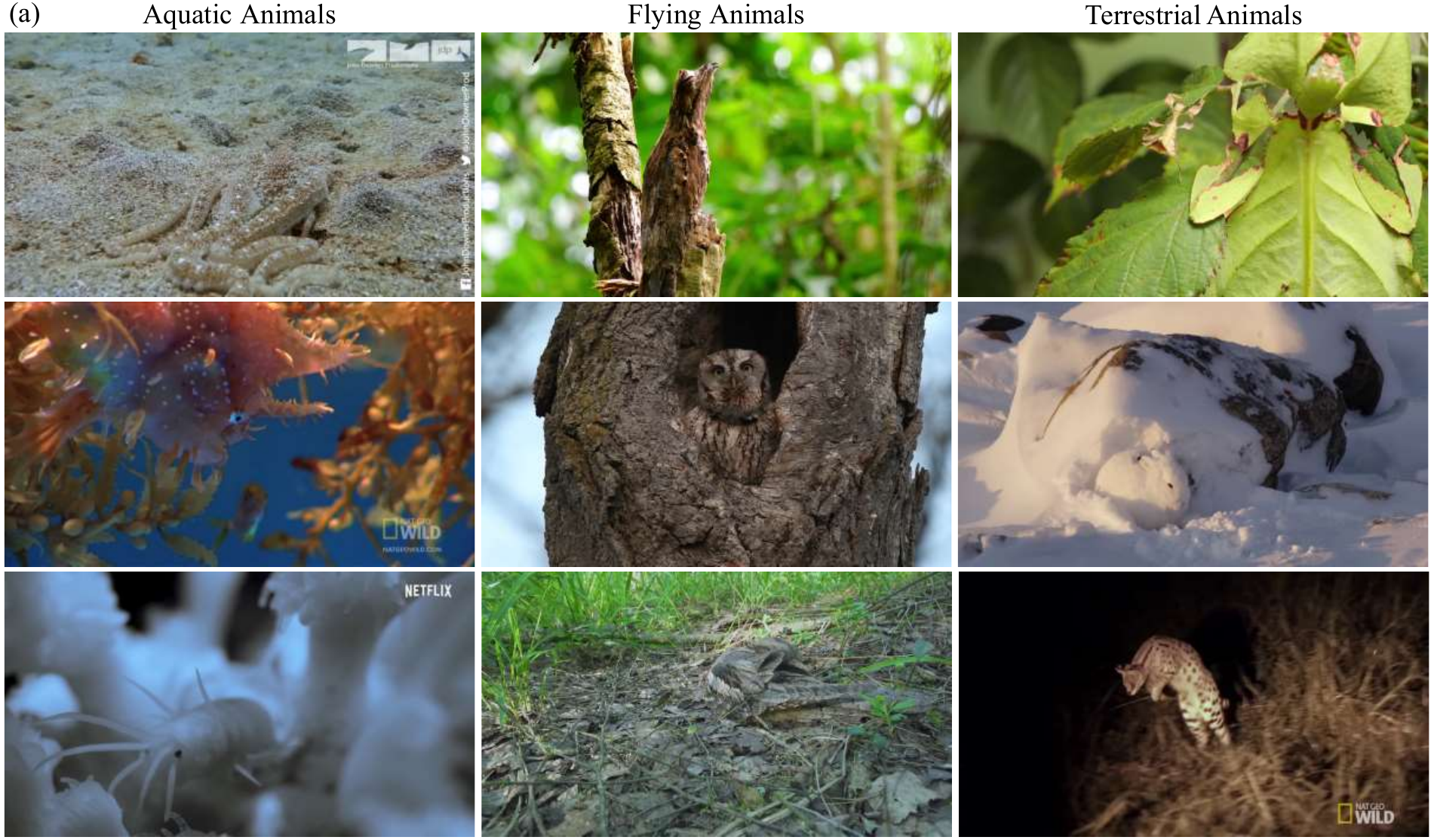}
    
    \vspace{0.5cm} 
    
    \includegraphics[width=\textwidth]{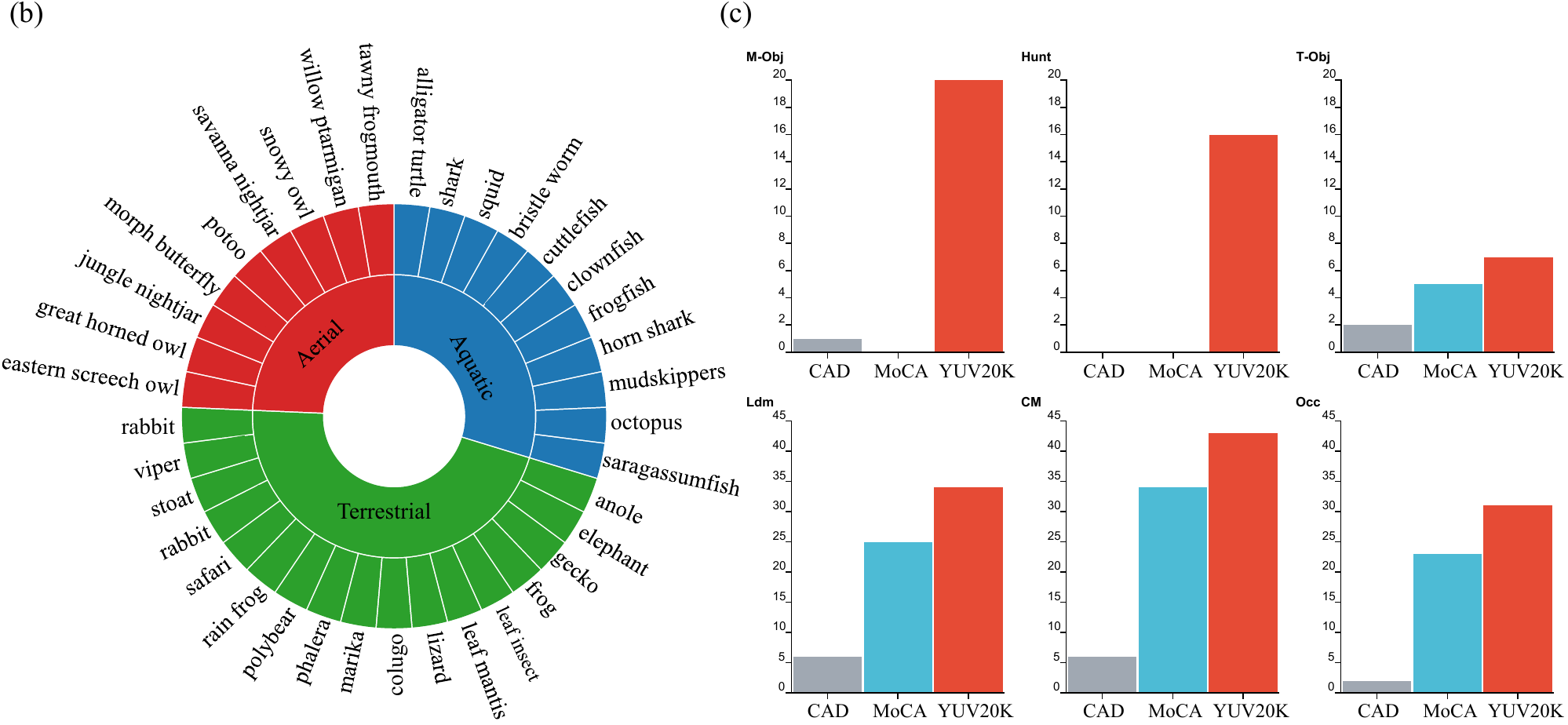}
    
    \caption{\textbf{Overview of the proposed YUV20K dataset.} (a) Representative frames showcasing diverse camouflage strategies across aquatic, aerial, and terrestrial environments. (b) A hierarchical sunburst chart illustrating the rich taxonomic diversity of the annotated animal species. (c) Attribute-based comparison with existing datasets (CAD and MoCA-Mask) across six complex scenarios: Large Displacement Motion (\textbf{Ldm}), Camera Motion (\textbf{CM}), Occlusion (\textbf{Occ}), Multiple Objects (\textbf{M-Obj}), Hunting (\textbf{Hunt}), and Tiny Object (\textbf{T-Obj}). YUV20K provides a significantly larger number of challenging videos across all attributes, notably establishing a new benchmark in extreme cases like M-Obj and Hunt, where previous datasets are severely lacking.}
    \label{fig:overview_dataset}
\end{figure*}

\section{YUV20K Dataset}

Large-scale, high-quality datasets serve as the cornerstone for advancing and rigorously evaluating detection algorithms. The rapid progress in Image COD (ICOD) is largely attributed to the availability of well-established benchmarks, such as CAMO\cite{le2019anabranchcamo}, CHAMELEON\cite{skurowski2018animalchameleon}, COD10K \cite{fan2021concealed}, and NC4K \cite{yunqiu_cod21}.

Conversely, the evolution of Video COD (VCOD) has been heavily bottlenecked by data scarcity. For instance, the pioneering VCOD dataset, MoCA \cite{lamdouar2020betrayed}, primarily provides bounding box annotations rather than dense pixel-level masks. Cheng et al. \cite{cheng2022implicit} subsequently introduced MoCA-Mask to supply pixel-level labels, the overall scene diversity and motion complexity remain limited. Similarly, another available dataset, CAD \cite{bideau2016scad}, contains only 9 short video sequences, which is far from sufficient for comprehensive model evaluation.

Compared to the abundance of ICOD data, the limited scale and lack of challenging scenarios in current VCOD datasets severely hinder the rigorous evaluation of spatiotemporal models\cite{xiao2024survey}. To bridge this critical gap and provide a robust testing ground for the community, we introduce the \textbf{YUV20K} dataset, featuring highly diverse distributed environments, complex scenarios, and high-quality frame-level annotations.

\subsection{Dataset Construction}
\textbf{Dataset Collection and Annotation}.

To ensure biological diversity and scene complexity, we curated raw videos from open-source platforms (e.g., YouTube, Google) using a diverse set of keywords covering 47 animal families. We strictly screened the videos based on three criteria: (1) High Camouflage Degree, excluding instances with obvious salient features; (2) Motion Richness, prioritizing clips containing erratic movements or deformations; and (3) High Resolution, ensuring sufficient detail for pixel-level annotation.
This process yielded 91 video clips, comprising a total of 24,295 frames with an average length of 266 frames per clip.

\textbf{Annotation Pipeline.}
Annotating camouflaged objects frame-by-frame is labor-intensive and error-prone. To balance efficiency and quality, we employed a semi-automatic pipeline. 
First, we utilized X-AnyLabeling\cite{X-AnyLabeling} integrated with state-of-the-art foundation models (SAM 2\cite{ravi2024sam} and Grounding SAM\cite{ren2024groundedgroundsam}) to generate initial coarse masks. Second, these proposals underwent a rigorous manual refinement phase to correct boundary errors, with special attention given to motion-blurred regions and intricate biological structures (e.g., feathers, limbs). Finally, a cross-validation round was conducted to resolve any remaining ambiguities.

\noindent\textbf{Diversity Explanation}
\textit{Category Diversity.}
As illustrated in Tab.~\ref{tab:dataset_comparison}, our dataset comprises 47 distinct animal species, providing a rich taxonomic diversity that spans a broad spectrum of ecological niches. This comprehensive coverage exposes models to a vast array of biological camouflage strategies, effectively preventing algorithms from overfitting to specific appearance priors and ensuring robust generalization across unseen species.

\textit{Habitat Diversity.}
As illustrated in Fig.~\ref{fig:overview_dataset} (a) and (b), we construct a wide range distribution of ecological habitats to accurately reflect real-world scenarios. The dataset systematically covers terrestrial, aquatic, and aerial domains, thereby ensuring a comprehensive representation of diverse background clutter, varying complex environmental structures.

\textit{Scenario Diversity.}
To faithfully reflect real-world complexities, YUV20K incorporates a wide spectrum of challenging spatiotemporal scenarios. As detailed in Fig.~\ref{fig:overview_dataset} (c), the dataset encompasses not only complex motion dynamics—such as camera motion, large displacements, and non-rigid deformations—but also diverse target attributes and behaviors, including hunting events, tiny-scale targets, and multi-object instances. This comprehensive profile provides a rigorous testbed for evaluating the robustness of VCOD algorithms in the wild.

\textit{Spatial Distribution Diversity}
A common limitation in video datasets is a noticeable central tendency, where subjects frequently appear near the center of the frame due to human filming habits. As illustrated in Fig~\ref{fig:density_comparison}, the spatial density maps reveal that YUV20K provides a substantially broader and more uniform spatial distribution. This extensive spatial coverage effectively mitigates center bias, requiring models to develop robust spatiotemporal segmentation capabilities.

\section{Method}

\subsection{Overall Architecture}
\begin{figure*}[t] 
    \centering
    \includegraphics[width=\textwidth]{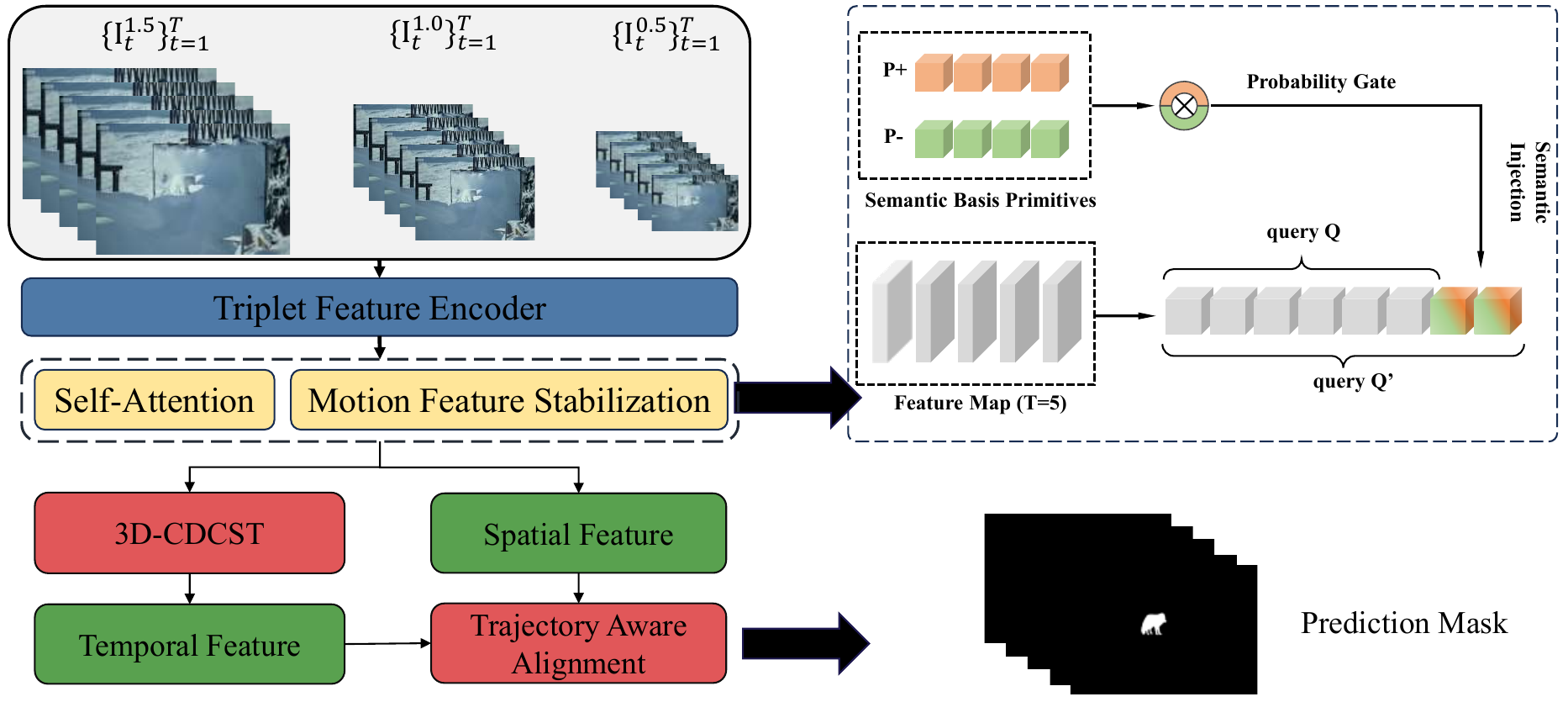}
    \caption{\textbf{Overall architecture of the proposed framework.} Given a video sequence, we construct a multi-scale input pyramid ($0.5\times, 1.0\times, 1.5\times$) and extract hierarchical features via a shared backbone. Then features are refined by self-attention with Motion Feature Stabilization module to capture global context. The next step decouples the representations into two pathways. In the temporal pathway, the 3D Central DIfference Spatial-Temporal Convolution modules extract robust temporal cues resilient to severe motion dynamics. In spatial way, the output of self-attention directly push into Trajectory-Aware Alignment module(TAA). Guided by these temporal features, the TAA module dynamically samples and aligns the spatial features. Finally, a decoder aggregates these aligned representations for accurate camouflage mask prediction.}
    \label{fig:overallstructure}
\end{figure*}

As illustrated in Fig.~\ref{fig:overallstructure}, our framework is designed to process video sequences and accurately segment camouflaged objects by overcoming motion-induced instability and temporal feature misalignment. Given a video clip consisting of 5 consecutive frames, denoted as $I = \left \{  I_t^s\right \}^T$, $s=0.5,1,1.5$. We first feed them into a Triplet Feature Encoder\cite{wang2022pvtv2} to extract hierarchical spatial features.

While these multi-scale features capture rich spatial details, they are fundamentally vulnerable to the complex motion dynamics inherent in real-world camouflaged scenarios. To tackle this, the extracted features are first processed by the self-attention with \textbf{Motion Feature Stabilization (MFS)} module. The MFS introduces robust semantic anchors to alleviate the appearance instability caused by motion or severe deformations, ensuring feature consistency within the spatial domain. The output feature is defined as the \textbf{spatial features}.
To further capture inter-frame motion dynamics, the stabilized features are fed into the \textbf{3DCDCST} module. The output of this module effectively encapsulates continuous motion cues and is defined as the \textbf{temporal features}.Finally, to resolve the issue of temporal misalignment, the \textbf{Trajectory-Aware Alignment (TAA)} module takes both features as input. It leverages the extracted temporal features to explicitly predict motion offsets, which are then used to perform deformable sampling on the spatial features, achieving precise spatiotemporal alignment. The aligned, multi-scale features are ultimately aggregated and fed into a decoder to generate the final camouflaged object masks. In the following subsections, we detail the design of these core components.

\subsection{Motion Feature Stabilization module}
\subsubsection{Semantic Basis Primitives}
Complex motion inevitably induces appearance instability (e.g., blur), causing significant disturbances in feature representations. Consequently, a stable reference or is requisite to rectify these features. To address this, we propose the Motion Feature Stabilization (MFS) module. 
Drawing inspiration from recent advances in fine-grained visual classification and confusing image recognition~\cite{espinosa2022concept,koh2020concept,ye_ika2:_2026}, we shift our attention to the intrinsic semantic attributes of camouflaged targets (e.g., biological textures like skin, feather), termed as Semantic Basis Primitives (SBPs). Unlike dynamic video features that fluctuate over time, these primitives function as anchors. They are injected into the feature stream to implicitly discover and aggregate generalized camouflage concepts. By utilizing these anchors to perform global query operations, we can leverage their stability to probe the motion-perturbed feature space and efficiently localize the camouflaged targets.

Technically, we instantiate the SBPs as a set of global learnable parameters, independent of the input video frames. To enhance the discriminative power specifically to distinguish subtle camouflaged patterns from complex backgrounds, we structure them into paired components. 
Let $\mathcal{P} = \{(\mathbf{p}_i^{+}, \mathbf{p}_i^{-})\}_{i=1}^{N}$ denote the set of $N$ primitive pairs, where $\mathbf{p}_i^{+} \in \mathbb{R}^C$ represents the \textit{positive} capturing target-related semantics, and $\mathbf{p}_i^{-} \in \mathbb{R}^C$ represents the \textit{negative} responsible for suppressing background semantics. We initialize these embeddings via a standard Gaussian distribution to ensure a diverse semantic starting point. 
This paired structure allows the module to explicitly model both the \textit{presence} of camouflaged attributes and the \textit{absence} of background distractors, forming a comprehensive concept basis.

\subsubsection{Stabilization Injection}
For each semantic primitive $i$, a learnable logit $l_i$ governs the mixing probability $\alpha_i = \sigma(l_i)$, where $\sigma$ represents for sigmoid operation. The final concept token $\mathbf{c}_i$ is synthesized via convex combination. This mechanism allows the module to adaptively emphasize or suppress specific camouflage patterns.

\begin{equation}
\mathbf{C} = \sum_{i}^{N} [\alpha_i \mathbf{p}_i^+ + (1 - \alpha_i) \mathbf{p}_i^- ]
\end{equation}.

Upon obtaining the stable concept tokens, we integrate them into the feature stream via a self-attention mechanism. Specifically, we inject these vectors into the feature map $F_s$ generated query $Q$ to formulate an augmented query set, denoted as $\mathbf{Q}'$, which can be seen in top-right of Fig.~\ref{fig:overallstructure}.  This augmented representation is subsequently linearly projected to generate the corresponding Keys ($\mathbf{K}$) and Values ($\mathbf{V}$). By equipping the query with these stable concept vectors, the attention mechanism gains the capability to globally screen motion-induced interference and actively probe for generalized camouflage attributes (e.g., texture patterns). Consequently, this interaction yields a rectified feature map, where the target representation is stabilized against dynamic perturbations.

\begin{equation}
    \mathbf{X} = Concat[\mathbf{F}_{s} , Linear(\mathbf{C})] ,
\end{equation}
\begin{equation}
    \mathbf{Q'} = \mathbf{X}\mathbf{W}_{Q'}, \quad \mathbf{K} = \mathbf{X}\mathbf{W}_K, \quad \mathbf{V} = \mathbf{X}\mathbf{W}_V,
\end{equation}
\begin{equation}
    F_{s} = \text{Softmax}\left(\frac{\mathbf{Q'}\mathbf{K}^\top}{\sqrt{d_k}}\right)\mathbf{V}.
\end{equation}

\subsection{Trajectory Aware Alignment}
\subsubsection{3D Central Difference Spatial-Temporal Convolution}

In VCOD scenarios, the efficacy of trajectory alignment heavily relies on the quality of motion guidance. However, due to the high visual homogeneity between the target and the background, standard feature extractors (e.g., vanilla 3D convolutions) that primarily aggregate intensity information often fail to capture reliable motion boundaries. Consequently, the subsequent offset prediction module (TAA) lacks reliable guidance, ultimately leading to severe feature misalignment.

To effectively guide trajectory alignment, it is crucial to extract fine-grained motion cues while suppressing static background textures—a task where standard intensity-based 3D convolutions often fail. To address this, we introduce the 3D Central Difference Spatial Temporal Convolution (3D-CDCST) \cite{yu2020searching3dcdc,yu2021dual}, which integrates gradient-level differential information into spatiotemporal aggregation. Specifically, we formulate 3D-CDCST as a hyperparameter-controlled fusion of a vanilla intensity term and a central difference gradient term. For efficient implementation, we derive it as a unified operator that shares weights without introducing extra parameters:

\begin{equation}
\begin{aligned}
\mathbf{y}(p_0) &= \theta \cdot \underbrace{\sum_{p_n} w(p_n)(x(p_0+p_n) - x(p_0))}_{\text{Gradient Term}} \\
&\quad + (1-\theta) \cdot \underbrace{\sum_{p_n} w(p_n)x(p_0+p_n)}_{\text{Vanilla Term}} \\
&= \underbrace{\sum_{p_n \in \mathcal{R}} w(p_n) \cdot x(p_0 + p_n)}_{\text{Standard 3D Conv}} \\
&\quad - \theta \cdot x(p_0) \cdot \underbrace{\sum_{p_n \in \mathcal{R}} w(p_n),}_{\text{spatio-temporal term}}
\end{aligned}
\end{equation}
where $p_0$ is the center position, $\mathcal{R}$ is the local neighborhood, and $\theta \in [0, 1]$ balances the contribution of gradient versus intensity. This design highlights motion boundaries, producing a high-fidelity motion proxy $\mathbf{F}_{t}$ to accurately guide the subsequent deformable grid sampling.

\begin{figure}[htbp]
    \centering
    \includegraphics[width=\linewidth]{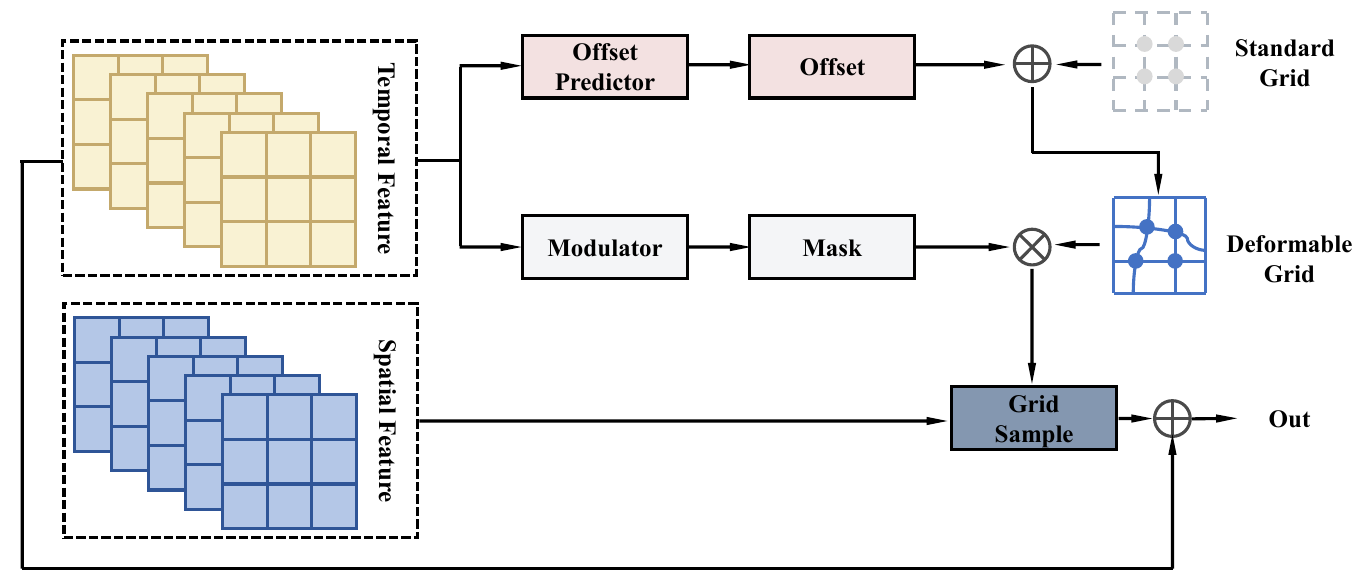}
    
    \caption{Detailed architecture of the Trajectory-Aware Alignment (TAA) module. The motion-enhanced temporal feature $\mathbf{F}_t$ is utilized to predict trajectory offsets $\Delta \mathcal{P}$ and a modulation mask $\mathcal{M}$. The offsets dynamically deform the standard sampling grid to accurately sample the static spatial feature $\mathbf{F}_s$. The aligned spatial representations are then modulated by $\mathcal{M}$ and add back into the temporal stream.}
    \label{fig:TAA}
\end{figure}
\subsubsection{Deformable Grid Sampling}
Standard sampling operations rely on fixed grids, which are suboptimal for capturing the complex, non-rigid motion patterns of camouflaged targets. To address this, we extend deformable convolution~\cite{dcnzhu2020deformabledetr,wang2023internimage} into a Trajectory-Aware Alignment (TAA) module, as illustrated in Fig.~\ref{fig:TAA}. The primary goal is to resolve the spatial misalignment between the static appearance features $\mathbf{F}_{s}$ and the instantaneous motion features $\mathbf{F}_{t}$.

\noindent\textbf{Trajectory Prediction.}
We utilize the motion-enhanced feature $\mathbf{F}_{t}$, which explicitly encodes motion boundaries as a guidance proxy. A lightweight 3D convolution layer $\phi$ is applied to $\mathbf{F}_{t}$ to simultaneously regress the trajectory offsets $\Delta \mathcal{P}$ and a modulation mask $\mathcal{M}$:

\begin{equation}
\Delta \mathcal{P} = \phi_{off} \left ( \mathbf{F}_{t} \right ), \quad \mathcal{M} = \phi_{mod} \left ( \mathbf{F}_{t} \right ) \in \mathbb{R}^{T\times H\times W }.
\end{equation}

Here, $\Delta \mathcal{P}$ models the non-rigid deformation field, while the mask $\mathcal{M}$ serves as a confidence map to assign lower weights to unreliable sampling points (e.g., background noise).

\noindent\textbf{Alignment and Fusion.}
Guided by the predicted offsets $\Delta \mathcal{P}$ and the sampling grid $\mathcal{G}$, we perform bilinear sampling $\mathcal{S}$ to warp the appearance features $\mathbf{F}_{s}$ to align with the target's current position. The aligned features are then modulated and fused back into the motion stream via a residual connection:

\begin{equation}
\mathbf{F}_{out} = \mathbf{F}_{t} + \text{Linear} \left( \mathcal{S} \big( \mathbf{F}_{s}, \ \mathcal{G} + \Delta \mathcal{P} \big) \odot \mathcal{M} \right),
\end{equation}
where $\mathcal{S}(\cdot)$ denotes the bilinear sampling operator. By employing the TAA module, the stable appearance attributes are effectively synchronized with the motion features, yielding a rectified representation that is robust to complex dynamic scenarios.

\section{Experiments}
\subsection{Experimental Setup}

\subsubsection{Datasets}
We comprehensively evaluate our proposed framework on three pixel-level annotated VCOD datasets: CAD\cite{bideau2016scad}, MoCA-Mask\cite{cheng2022implicit}, and our YUV20K. 

The Camouflaged Animal Dataset (CAD) comprises 9 brief video sequences (839 frames) and serves exclusively as a test set. Built upon the original MoCA dataset\cite{lamdouar2020betrayed}, the MoCA-Mask\cite{cheng2022implicit} benchmark provides precise pixel-level annotations. Following the standard split in \cite{cheng2022implicit}, it is partitioned into a training set of 71 sequences (19,313 frames) and a test set of 16 sequences (3,626 frames). Furthermore, our proposed YUV20K contains 91 video clips (24,295 frames) covering 47 animal species, which is carefully divided into a training set of X sequences (Y frames) and a test set of Z sequences (W frames).

To rigorously assess both in-domain performance and cross-domain generalization, we establish two distinct training paradigms. 
\textbf{First}, when optimized on the MoCA-Mask training set, the model is evaluated in-domain on the MoCA-Mask test set, and cross-domain on CAD and the \textit{entirety} of the YUV20K dataset (combining its training and testing splits). 
\textbf{Second}, conversely, when trained on the YUV20K training set, the model is evaluated in-domain on the YUV20K test set, and cross-domain on CAD and the \textit{entirety} of the MoCA-Mask dataset (all training and testing splits).

\subsubsection{Evaluation Metrics}
To quantitatively evaluate the segmentation performance, we adopt seven widely used metrics implemented via the PySODMetrics library~\cite{pang2020pysodmetrics}. These include Structure-measure ($S_m$)~\cite{Smeasure}, maximum F-measure ($F_m$)~\cite{achanta2009frequencyfmeasure}, weighted F-measure ($F^w_\beta$)~\cite{wFmeasure}, Enhanced-alignment measure ($E_m$)~\cite{Emeasure}, and Mean Absolute Error ($M$)~\cite{MAE}. Additionally, we report the mean Dice coefficient (mDice) and mean Intersection over Union (mIoU) to provide a comprehensive assessment of the region-level segmentation accuracy.

\subsubsection{Implementation Details}
Our proposed framework is implemented in PyTorch. The spatial encoder is initialized with the parameters of PVTv2~\cite{wang2022pvtv2} pretrained on ImageNet. The network is optimized using the AdamW optimizer ($\beta_1=0.9, \beta_2=0.999$, weight decay=$1\times 10^{-6}$) with an initial learning rate of $1 \times 10^{-4}$, which is regulated by a cosine annealing learning rate scheduler. 

To ensure a fair comparison and facilitate our cross-domain evaluation, we follow the widely adopted two-stage training protocol~\cite{cheng2022implicit}. Specifically, the model is first pretrained on the static image camouflage dataset (COD10K-TR) to acquire fundamental appearance-based camouflage representations. Subsequently, to model complex spatiotemporal dynamics, we conduct two independent fine-tuning branches: one optimized exclusively on the MoCA-Mask training set, and the other solely on our YUV20K training set. Both models are fine-tuned for 10 epochs.

\subsection{Comparison with State-of-the-Art Methods}
To rigorously evaluate the effectiveness of our proposed framework, we conduct a comprehensive comparison with \textbf{7} recent state-of-the-art (SOTA) methods. These encompass both prominent image-based models (e.g., ZoomNet~\cite{pang2022zoom}), and recent strong video-based baselines (e.g.,ZoomNext~\cite{pang2024zoomnext}, EMIP~\cite{zhang2025explicitemip}). To ensure absolute fairness, all comparison methods are evaluated using their official codebases and provided weights, or retrained strictly under our identical experimental protocols.

Following our dual-source training paradigm, the quantitative results are reported in two distinct settings:
\textbf{First}, Tab.~\ref{tab:train_on_moca} summarizes the performance of models optimized on the MoCA-Mask training set. In this setting, alongside Video COD models, we purposefully include top-tier Image COD methods as static baselines. It is worth noting that these image-based models are evaluated directly using their officially released static weights without any video-specific fine-tuning, serving as a reference for pure appearance-based detection capabilities. 

\textbf{Second}, Tab.~\ref{tab:train_on_yuv} presents the results of models trained on our proposed YUV20K dataset. Since YUV20K is specifically designed to benchmark highly complex scenarios, this table focuses exclusively on comparing dedicated Video COD architectures. This allows for a rigorous assessment of each model's true spatiotemporal learning capacity when confronted with extreme non-rigid deformations and large displacements.

\begin{figure*}[t]
    \centering
    \includegraphics[width=\textwidth]{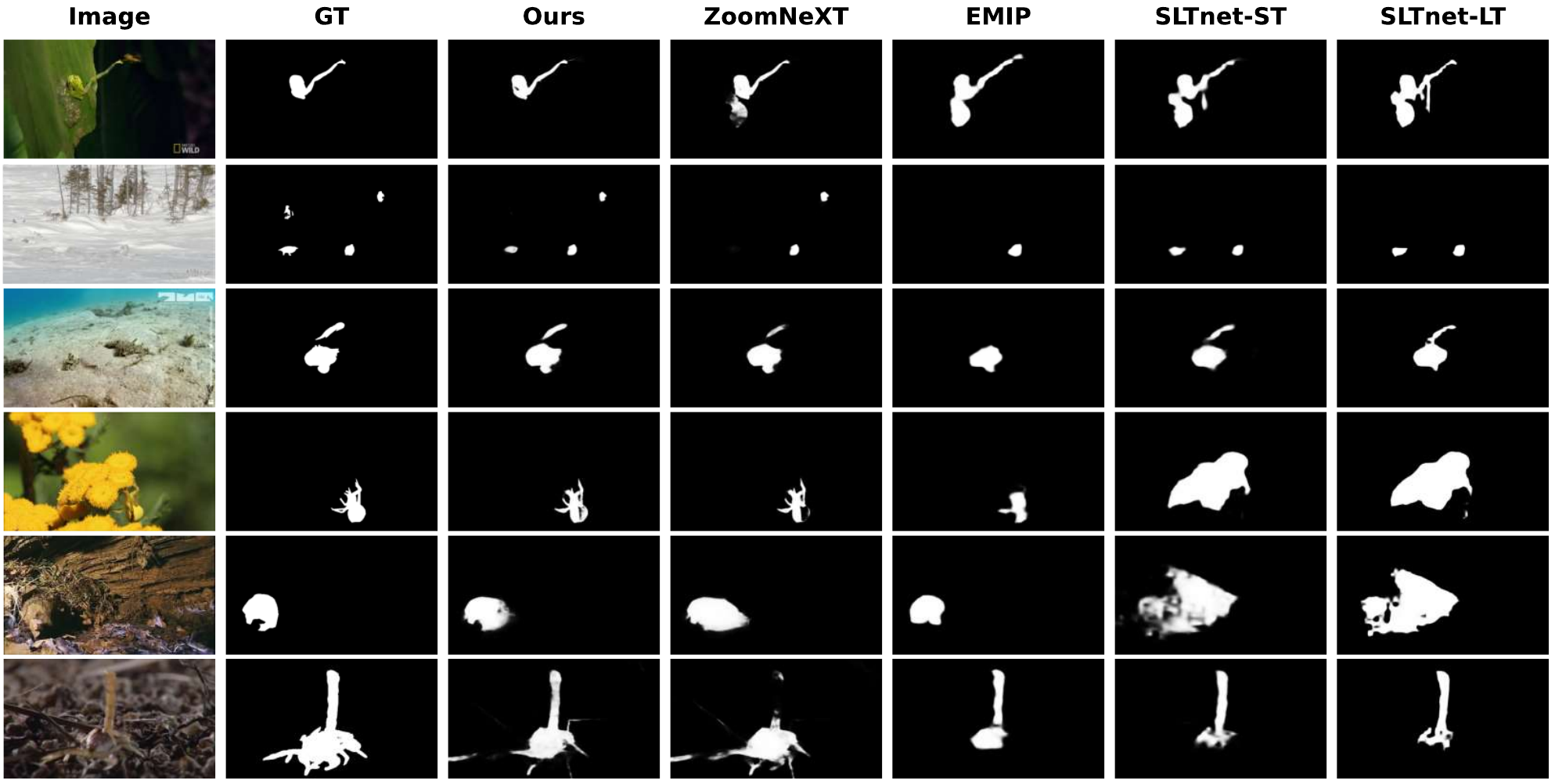}
    \caption{
        \textbf{Qualitative visual comparison of our proposed approach against state-of-the-art methods.} 
        Representative sequences are selected from the YUV20K MoCA, and CAD datasets (top to bottom). 
        From left to right: the original input frames, ground truth (GT) masks, predictions of our model, ZoomNeXT, EMIP, and SLT-Net (Short-term and Long-term variants). 
    }
    \label{fig:qualitative_comparison}
\end{figure*}

Overall, our method consistently achieves superior performance across both training paradigms. Finally, qualitative visual comparisons against competitive baselines are presented in Fig.~\ref{fig:qualitative_comparison} and Fig.~\ref{fig:vis_seq_comparison}, further demonstrating our model's robustness in effectively suppressing motion feature instability and accurately aligning moving camouflaged targets.

\subsubsection{Quantitative Evaluation}

Tab.~\ref{tab:train_on_moca} and Tab.~\ref{tab:train_on_yuv} report the quantitative performance of all competing methods across three benchmarks: CAD, MoCA-Mask, and our proposed YUV20K. 

As observed, our framework consistently sets new state-of-the-art records across both training paradigms, though the performance dynamics vary to reflect the intrinsic complexity of the source domains. Specifically, when optimized on the MoCA-Mask dataset (Tab.~\ref{tab:train_on_moca}), our method significantly surpasses recent strong competitors, such as ZoomNext~\cite{pang2024zoomnext} and EMIP~\cite{zhang2025explicitemip}, by a notable margin across most metrics.

Conversely, training on our proposed YUV20K benchmark (Tab.~\ref{tab:train_on_yuv}) presents a formidable challenge due to its highly complex scenarios. This unprecedented difficulty inherently narrows the performance margins among top-tier methods. Furthermore, it is noteworthy that most models achieve relatively high absolute metric scores on the YUV20K test set. We primarily attribute this favorable performance to the high-definition (1080p) nature of our dataset. Unlike earlier benchmarks plagued by low-resolution compression artifacts, the 1080p sequences in YUV20K inherently preserve fine-grained texture cues and delicate boundary details. This high spatial fidelity equips all models with richer foundational signals, thereby raising the overall performance baseline.

Beyond raising the evaluation baseline, this high spatial fidelity also makes YUV20K an exceptionally superior training data source. A compelling testament to this is the remarkable cross-domain performance of EMIP on MoCA-Mask (e.g., achieving an $S_\alpha$ of 0.879) despite being trained exclusively on YUV20K. The rich, noise-free spatiotemporal details in our 1080p sequences empower optical-flow-based methods to extract highly accurate motion boundaries and learn explicit motion priors to their maximum potential. This demonstrates that YUV20K serves as a powerful foundational resource that significantly enhances the representation capabilities of existing VCOD architectures.

However, while our high-quality data maximizes the potential of these baselines, it simultaneously exposes their inherent architectural limitations. Even with strong priors acquired from YUV20K, EMIP exhibits a fundamental trade-off. By architecturally enforcing a strict reliance on explicit motion, EMIP perfectly aligns with the distinct, large-scale movements in MoCA-Mask but suffers significant degradation in subtle or static-like camouflage scenarios, as evidenced by its sharp performance drop on the cross-domain CAD dataset. Conversely, our framework avoids these aggressive motion shortcuts. Our explicit trajectory-aware alignment (TAA) and motion feature stabilization (MFS) mechanisms prioritize global semantic stability. Consequently, we trade negligible in-domain structural gains for significantly superior cross-domain robustness and exceptionally clean background suppression, achieving a leading Mean Absolute Error ($\mathcal{M}$ of 0.015) on YUV20K. This confirms that our model is optimized for balanced, real-world generalization rather than overfitting to specific motion distributions.

\begin{table*}[t]
    \centering
    \caption{\textbf{Quantitative comparison of models trained on MoCA-Mask.} All models in this table are trained exclusively on the MoCA-Mask training set and evaluated on the test sets of CAD (cross-domain), MoCA-Mask (in-domain), and our proposed YUV20K (cross-domain). \textbf{Bold} and \textcolor{red}{\textbf{red}} indicate the best performance. $\uparrow$ denotes higher is better, and $\downarrow$ denotes lower is better. The symbol ``-'' indicates the results are not available.}
    \label{tab:train_on_moca}

    \setlength{\tabcolsep}{2.5pt}
    
    \resizebox{\textwidth}{!}{
        \begin{tabular}{ l | ccccccc | ccccccc | ccccccc }
            \toprule
            \multirow{2}{*}{\textbf{Method}}  & 
            \multicolumn{7}{c|}{\textbf{CAD} \cite{bideau2016scad}} & 
            \multicolumn{7}{c|}{\textbf{MoCA-Mask} \cite{fan2020camouflaged}} &
            \multicolumn{7}{c}{\textbf{YUV20K} (Ours)} \\
            \cmidrule(lr){2-8} \cmidrule(lr){9-15} \cmidrule(lr){16-22}
            
             & $S_\alpha$$\uparrow$ & $F_\beta$$\uparrow$ & $F_\beta^\omega$$\uparrow$ & $\mathcal{M}$$\downarrow$ & $E_m$$\uparrow$ & mIoU$\uparrow$ & mDice$\uparrow$
             & $S_\alpha$$\uparrow$ & $F_\beta$$\uparrow$ & $F_\beta^\omega$$\uparrow$ & $\mathcal{M}$$\downarrow$ & $E_m$$\uparrow$ & mIoU$\uparrow$ & mDice$\uparrow$ 
             & $S_\alpha$$\uparrow$ & $F_\beta$$\uparrow$ & $F_\beta^\omega$$\uparrow$ & $\mathcal{M}$$\downarrow$ & $E_m$$\uparrow$ & mIoU$\uparrow$ & mDice$\uparrow$ \\
            \midrule
            
            \textit{Image-based Methods} & \multicolumn{21}{c}{} \\
            \midrule
            
            SINet \cite{fan2020camouflaged} 
            & .621 & .405 & .380 & .045 & .720 & .310 & .401  
            & .580 & .250 & .230 & .040 & .650 & .210 & .280 
            & - & - & - & - & - & - & - \\ 

            ZoomNet \cite{pang2022zoom} 
            & .652 & .430 & .410 & .040 & .750 & .340 & .440  
            & .600 & .280 & .250 & .035 & .670 & .230 & .310 
            & .775 & .645 & .599 & .039 & .791 & .535 & .615 \\

            \midrule
            \textit{Video-based Methods} & \multicolumn{21}{c}{} \\
            \midrule

            SLT-Net-ST \cite{cheng2022implicit} 
            & .719 & .517 & .479 & .032 & .832 & .410 & .509  
            & .631 & .311 & .288 & .029 & .701 & .258 & .344 
            & .842 & .740 & .712 & .031 & .888 & .652 & .738 \\ 

            SLT-Net-LT \cite{cheng2022implicit} 
            & .720 & .530 & .498 & .030 & .849 & .424 & .521  
            & .621 & .311 & .293 & .029 & .722 & .259 & .346 
            & .839 & .750 & .724 & .029 & .898 & .664 & .748 \\  
            
            IMEX \cite{hui2024implicit} 
            & .684 & - & .452 & .033 & .813 & .370 & .469
            & .661 & - & .371 & .020 & .778 & .409 & .319 
            & - & - & - & - & - & - & - \\

            EMIP \cite{zhang2025explicitemip} 
            & .710 & .534 & .504 & .029 & .835 & .415 & .528
            & .669 & .400 & .374 & .017 & .789 & .326 & .424 
            & .824 & .725 & .690 & .033 & .890 & .621 & .715 \\

            ZoomNext \cite{pang2024zoomnext}  
            & .759 & .613 & .562 & .020 & .861 & .485 & .565
            & .699 & .445 & .421 & .008 & .696 & .367 & .434 
            & .823 & .720 & .691 & .036 & .851 & .628 & .698 \\
            
            \midrule
            
            \textbf{Ours} 
            & \color{red}\textbf{.771} & \color{red}\textbf{.613} & \color{red}\textbf{.584} & \color{red}\textbf{.019} & \color{red}\textbf{.871} & \color{red}\textbf{.507} & \color{red}\textbf{.589}
            & \color{red}\textbf{.700} & \color{red}\textbf{.456} & \color{red}\textbf{.432} & \color{red}\textbf{.008} & \color{red}\textbf{.733} & \color{red}\textbf{.377} & \color{red}\textbf{.449} 
            & \color{red}\textbf{.858} & \color{red}\textbf{.812} & \color{red}\textbf{.780} & \color{red}\textbf{.035} & \color{red}\textbf{.920} & \color{red}\textbf{.697} & \color{red}\textbf{.772} \\
            \bottomrule
        \end{tabular}
    }
\end{table*}

\begin{table*}[t]
    \centering
    \caption{\textbf{Quantitative comparison of models trained on our proposed YUV20K.} All models in this table are trained exclusively on the YUV20K training set and evaluated on the test sets of CAD (cross-domain), MoCA-Mask (cross-domain), and YUV20K (in-domain). This demonstrates the strong generalization ability acquired from our dataset. \textbf{Bold} and \textcolor{red}{\textbf{red}} indicate the best performance. $\uparrow$ denotes higher is better, and $\downarrow$ denotes lower is better.}
    \label{tab:train_on_yuv}
    
    \setlength{\tabcolsep}{2.5pt}
    
    \resizebox{\textwidth}{!}{
        \begin{tabular}{ l | ccccccc | ccccccc | ccccccc }
            \toprule
            \multirow{2}{*}{\textbf{Method}}  & 
            \multicolumn{7}{c|}{\textbf{CAD} \cite{bideau2016scad}} & 
            \multicolumn{7}{c|}{\textbf{MoCA-Mask} \cite{fan2020camouflaged}} &
            \multicolumn{7}{c}{\textbf{YUV20K} (Ours)} \\
            \cmidrule(lr){2-8} \cmidrule(lr){9-15} \cmidrule(lr){16-22}
            
             & $S_\alpha$$\uparrow$ & $F_\beta$$\uparrow$ & $F_\beta^\omega$$\uparrow$ & $\mathcal{M}$$\downarrow$ & $E_m$$\uparrow$ & mIoU$\uparrow$ & mDice$\uparrow$
             & $S_\alpha$$\uparrow$ & $F_\beta$$\uparrow$ & $F_\beta^\omega$$\uparrow$ & $\mathcal{M}$$\downarrow$ & $E_m$$\uparrow$ & mIoU$\uparrow$ & mDice$\uparrow$ 
             & $S_\alpha$$\uparrow$ & $F_\beta$$\uparrow$ & $F_\beta^\omega$$\uparrow$ & $\mathcal{M}$$\downarrow$ & $E_m$$\uparrow$ & mIoU$\uparrow$ & mDice$\uparrow$ \\
            \midrule
            
            SLT-Net-ST \cite{cheng2022implicit} 
            & .720 & .492 & .445 & .039 & .806 & .398 & .499  
            & .798 & .642 & .597 & .029 & .848 & .559 & .659 
            & .848 & .684 & .648 & .025 & .908 & .607 & .697 \\ 
             
            EMIP \cite{zhang2025explicitemip} 
            & .725 & .533 & .483 & .030 & .839 & .405 & .518  
            & \color{red}\textbf{.879} & \color{red}\textbf{.789} & \color{red}\textbf{.775} & \color{red}\textbf{.012} & \color{red}\textbf{.925} & \color{red}\textbf{.722} & \color{red}\textbf{.789} 
            & .857 &\color{red}\textbf{.727} & \color{red}\textbf{.702} & .020 & \color{red}\textbf{.924} & \color{red}\textbf{.643} & .\color{red}\textbf{.717} \\

            ZoomNext \cite{pang2024zoomnext} 
            & \textbf{.775} & .569 & .565 & .018 & .815 & .503 & .577
            & .827 & .719 & .685 & .018 & .853 & .624 & .707 
            & .856 & .704 & .685 & .017 & .895 & .635 & .690 \\
            
            \midrule
            
            \textbf{Ours} 
            & \color{red}\textbf{.775} & \color{red}\textbf{.582} & \color{red}\textbf{.569} &  \color{red}\textbf{.017} &  \color{red}\textbf{.830} & \color{red}\textbf{.505} & 
            \color{red}\textbf{.578}
            & .831 & .726 & .693 & .017 & .857 & .632 & .714
            & \color{red}\textbf{.858} & .706 & .688 & \color{red}\textbf{.015} & .897 & .639 & .692 \\
            \bottomrule
        \end{tabular}
    }
\end{table*}

\begin{table*}[t]
    \centering
    \caption{\textbf{Attribute-based performance on the proposed YUV20K dataset.} We evaluate the video-based methods across six challenging motion scenarios: \textbf{Ldm} (Large displacement motion), \textbf{CM} (Camera Motion), \textbf{Occ} (Occlusion), \textbf{M-Obj} (Multiple Objects), \textbf{Hunt} (Hunting), and \textbf{T-Obj} (Tiny Object). We report the structure-measure ($S_\alpha$$\uparrow$), weighted F-measure ($F_\beta^\omega$$\uparrow$), and mean Intersection-over-Union ($mIoU$$\uparrow$) to demonstrate the robustness of different methods. \textbf{Bold} and \textcolor{red}{\textbf{red}} indicate the best performance.}
    \label{tab:attribute_comparison}
    
    \resizebox{\textwidth}{!}{
        \begin{tabular}{ l | ccc | ccc | ccc | ccc | ccc | ccc }
            \toprule
            \multirow{2}{*}{\textbf{Method}} & 
            \multicolumn{3}{c|}{\textbf{Ldm}} & 
            \multicolumn{3}{c|}{\textbf{CM}} & 
            \multicolumn{3}{c|}{\textbf{Occ}} & 
            \multicolumn{3}{c|}{\textbf{M-Obj}} & 
            \multicolumn{3}{c|}{\textbf{Hunt}} & 
            \multicolumn{3}{c}{\textbf{T-Obj}} \\
            \cmidrule(lr){2-4} \cmidrule(lr){5-7} \cmidrule(lr){8-10} \cmidrule(lr){11-13} \cmidrule(lr){14-16} \cmidrule(lr){17-19}
            
             & $S_\alpha$$\uparrow$ & $F_\beta^\omega$$\uparrow$ & $mIoU$$\uparrow$ 
             & $S_\alpha$$\uparrow$ & $F_\beta^\omega$$\uparrow$ & $mIoU$$\uparrow$ 
             & $S_\alpha$$\uparrow$ & $F_\beta^\omega$$\uparrow$ & $mIoU$$\uparrow$ 
             & $S_\alpha$$\uparrow$ & $F_\beta^\omega$$\uparrow$ & $mIoU$$\uparrow$ 
             & $S_\alpha$$\uparrow$ & $F_\beta^\omega$$\uparrow$ & $mIoU$$\uparrow$ 
             & $S_\alpha$$\uparrow$ & $F_\beta^\omega$$\uparrow$ & $mIoU$$\uparrow$ \\
            \midrule

            SLT-Net-ST \cite{cheng2022implicit} 
            & .855 & .745 & .689  
            & .885 & .802 & .747  
            & .877 & .784 & .716  
            & .733 & .542 & .474  
            & .782 & .624 & .547  
            & .639 & .364 & .307 \\ 
            
            SLT-Net-LT \cite{cheng2022implicit} 
            & .851 & .754 & .698  
            & .880 & .809 & .754  
            & .875 & .801 & .731  
            & .728 & .546 & .477  
            & .779 & .638 & .559  
            & .638 & .364 & .306 \\ 
            
            ZoomNet \cite{pang2022zoom} 
            & .799 & .644 & .583  
            & .826 & .695 & .636  
            & .791 & .648 & .573  
            & .678 & .428 & .358  
            & .726 & .525 & .440  
            & .637 & .281 & .235 \\ 

            ZoomNext \cite{pang2024zoomnext} 
            & .841 & .738 & .675  
            & .876 & .793 & .734  
            & \textbf{.901} & .842 & .768  
            & .773 & .611 & .539  
            & .824 & .699 & .624  
            & .691 & .426 & .366 \\ 
            
            EMIP \cite{zhang2025explicitemip} 
            & .832 & .709 & .647  
            & .859 & .758 & .695  
            & .863 & .768 & .689  
            & .721 & .520 & .447  
            & .736 & .552 & .468  
            & .629 & .321 & .271 \\ 
            
            \midrule
            
            \textbf{Ours} 
            & \textcolor{red}{\textbf{.856}} & \textcolor{red}{\textbf{.765}} & \textcolor{red}{\textbf{.700}}  
            & \textcolor{red}{\textbf{.886}} & \textcolor{red}{\textbf{.816}} & \textcolor{red}{\textbf{.755}}  
            & \textcolor{red}{\textbf{.901}} & \textcolor{red}{\textbf{.846}} & \textcolor{red}{\textbf{.769}}  
            & \textcolor{red}{\textbf{.780}} & \textcolor{red}{\textbf{.631}} & \textcolor{red}{\textbf{.553}}  
            & \textcolor{red}{\textbf{.825}} & \textcolor{red}{\textbf{.706}} & \textcolor{red}{\textbf{.627}}  
            & \textcolor{red}{\textbf{.695}} & \textcolor{red}{\textbf{.446}} & \textcolor{red}{\textbf{.372}} \\ 
            
            \bottomrule
        \end{tabular}
    }
\end{table*}

\begin{figure*}[t!p]
    \centering
    \includegraphics[width=\textwidth]{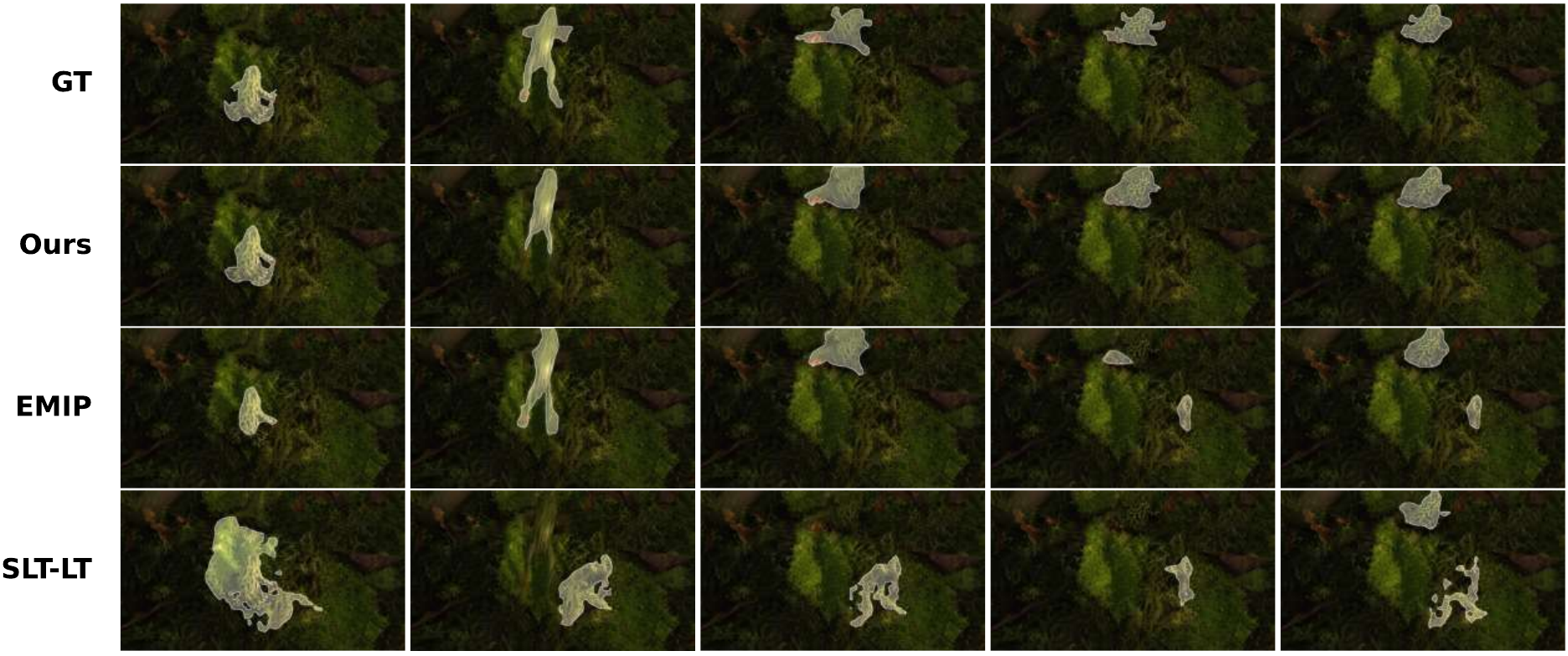}
    \vspace{0.1cm} 
    
    \includegraphics[width=\textwidth]{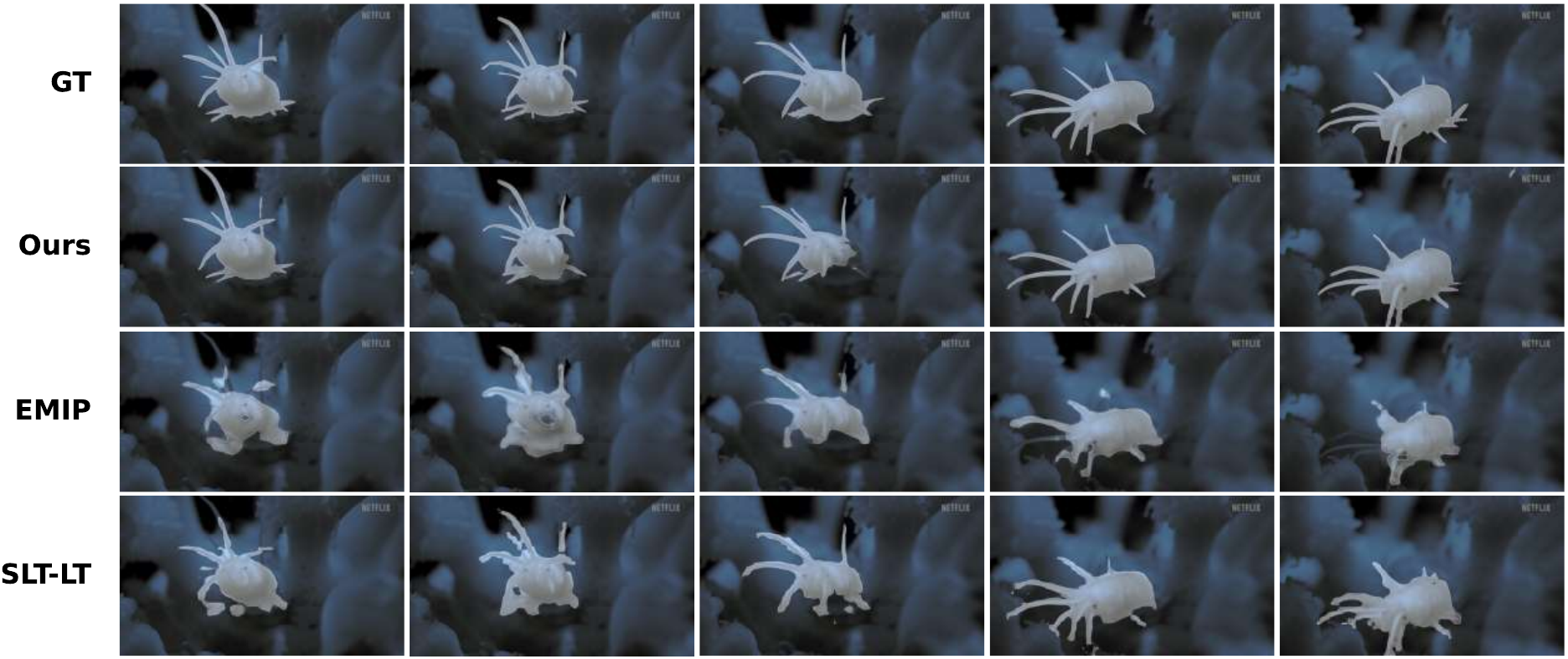}
    \vspace{0.1cm} 
    
    \includegraphics[width=\textwidth]{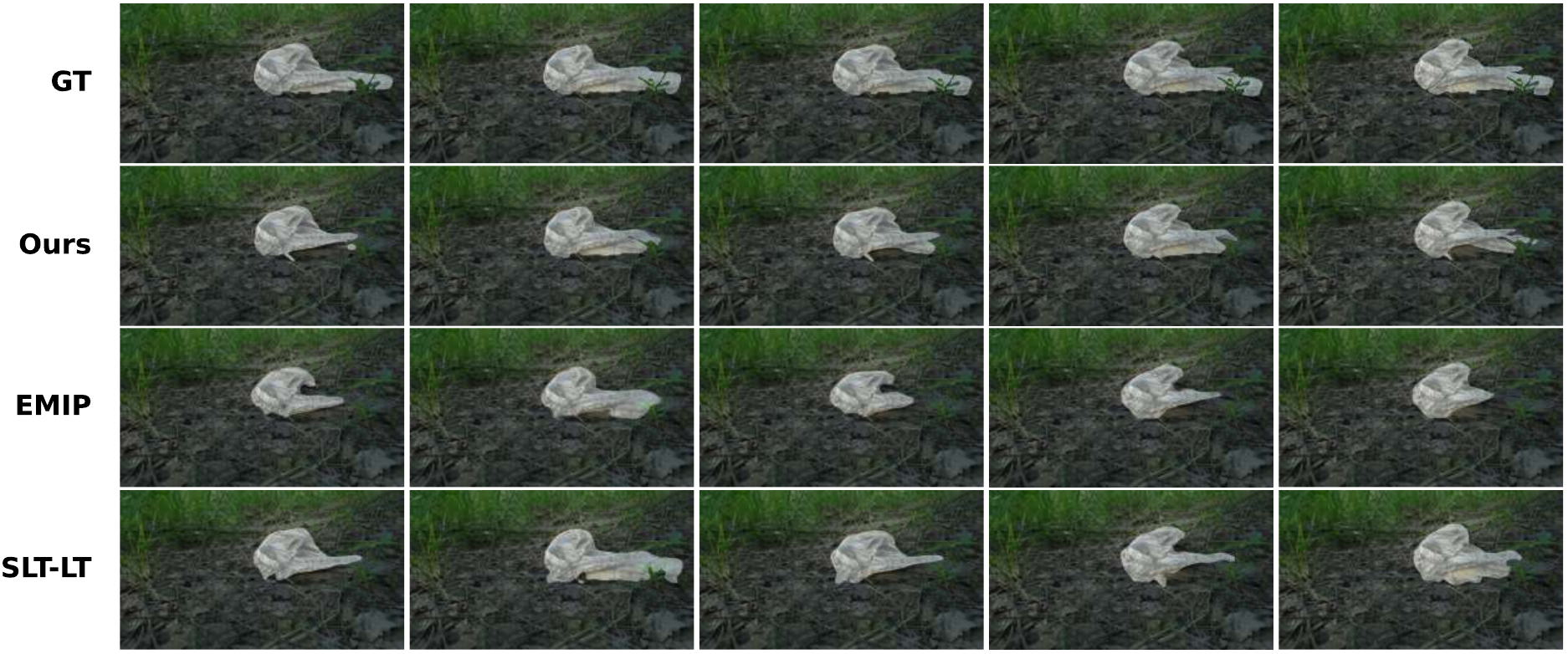}
    
    \caption{\textbf{Qualitative comparison of video camouflaged object detection.} Visual sequences comparing our proposed method against state-of-the-art models (e.g., EMIP, SLT-LT) on challenging scenarios from the YUV20K dataset (Part 1). By actively avoiding background interference through our deformable strategy, our method consistently maintains temporal coherence, predicts sharper boundaries, and effectively suppresses background noise compared to other approaches.}
    \label{fig:vis_seq_comparison}
\end{figure*}

\begin{figure*}[t!p]
    \ContinuedFloat 
    \centering
    \includegraphics[width=\textwidth]{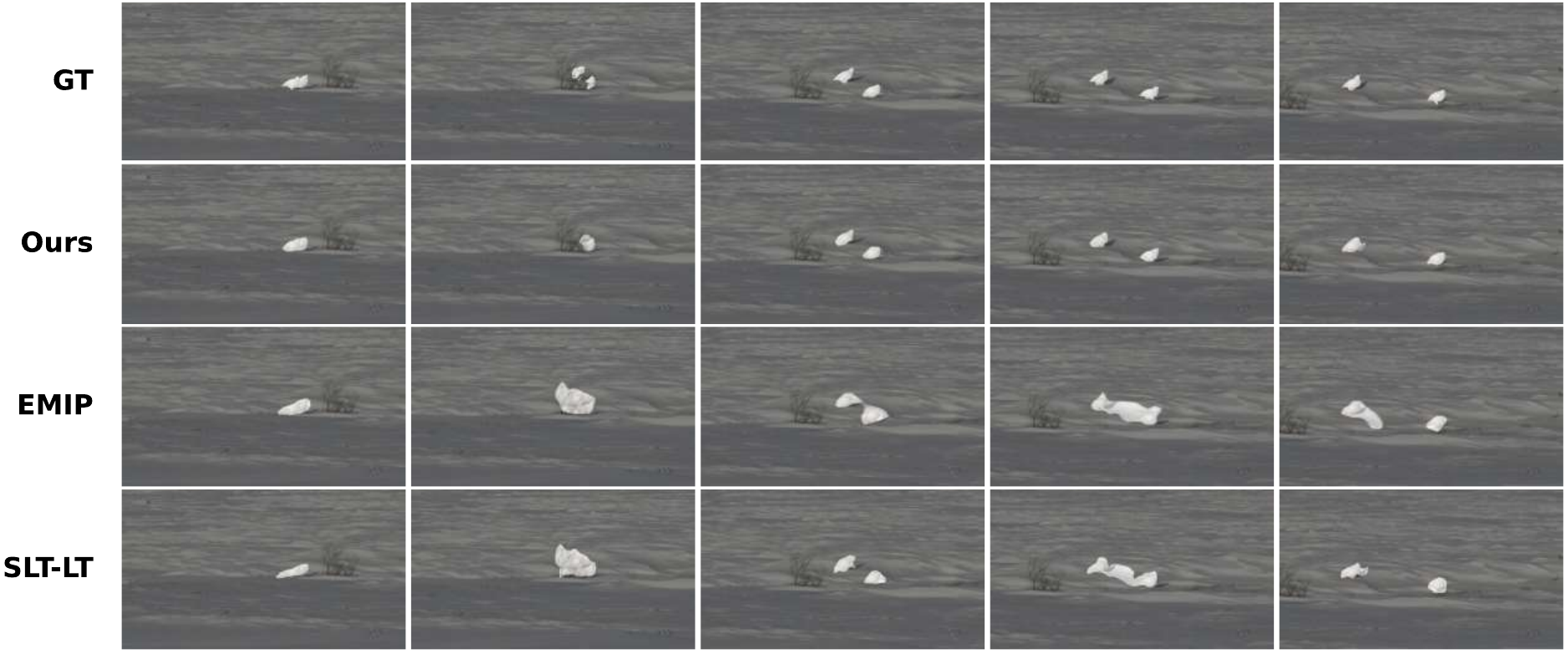}
    \vspace{0.1cm}
    
    \includegraphics[width=\textwidth]{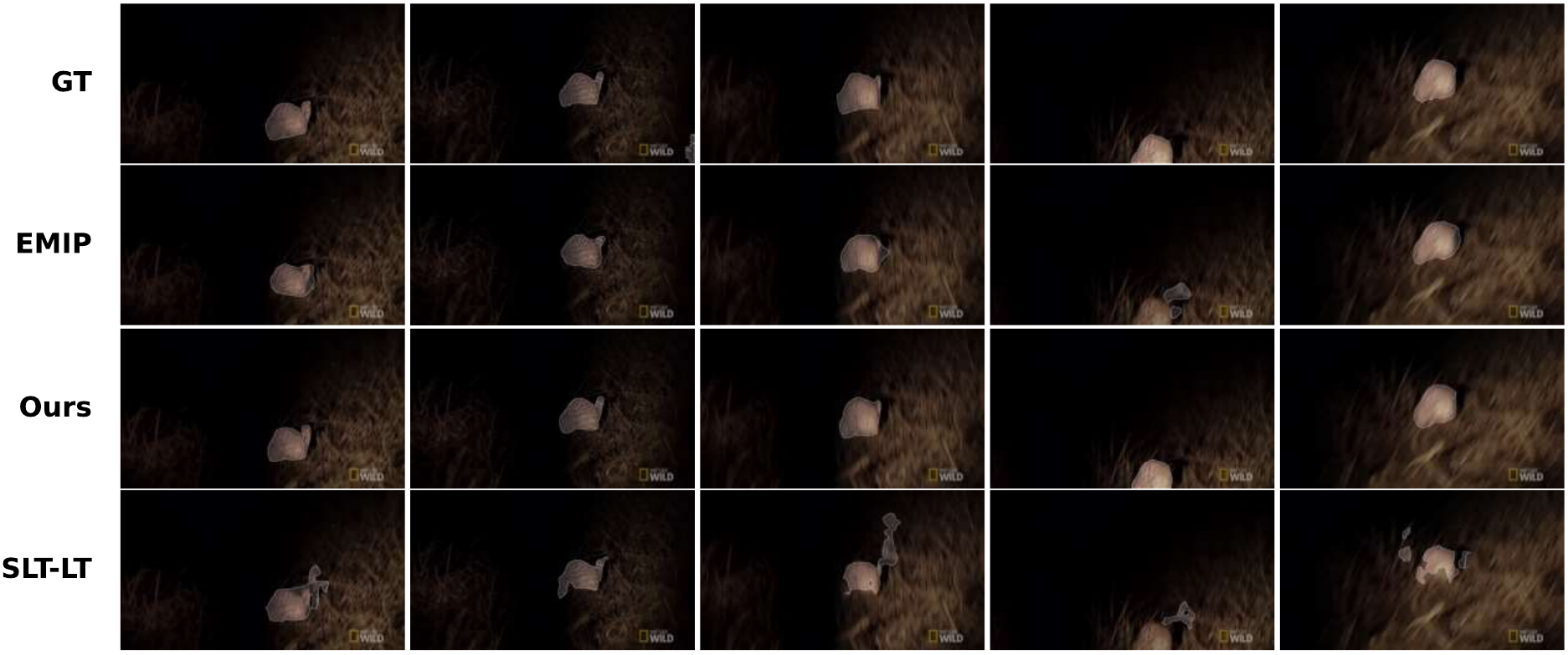}
    
    \caption{\textbf{Qualitative comparison of video camouflaged object detection (Continued).} Additional sequence results under extreme challenges such as heavy occlusion and tiny objects. Our model demonstrates superior robustness in faithfully tracking non-rigid targets across frames.}
\end{figure*}

\subsubsection{Qualitative Evaluation}
Visual comparison of with recent SOTA methods are shown in Fig.~\ref{fig:qualitative_comparison} and Fig.~\ref{fig:vis_seq_comparison}, which demonstrate the robustness of our model across various complex scenarios.
Specifically, Fig.~\ref{fig:qualitative_comparison} presents a comprehensive visual comparison across the three benchmarks, with each row highlighting a distinct and demanding challenge in video camouflaged object detection. 
The top three rows are sampled from our complex \textbf{YUV20K} dataset: the first row (\textit{frog}) illustrates sudden and rapid movements; the second row (\textit{birds in snow}) presents the extreme difficulty of identifying multiple, tiny targets; and the third row (\textit{flounder}) showcases severe non-rigid deformable motion. 
The subsequent two rows are selected from the \textbf{MoCA-Mask} benchmark: the fourth row (\textit{flower spider}) features subtle partial body motion where only specific part of spider are active; and the fifth row (\textit{hedgehog}) depicts extreme camouflage where the foreground heavily blends into the background. 
Finally, the last row showcases an example from the \textbf{CAD} dataset (\textit{scorpion}), introducing targets with highly complex topological shapes.

All methods suffer from severe motion-induced feature instability and temporal feature mismatch which often resulting in missing parts, blurred boundaries, or false positives under these demanding conditions. Our approach consistently maintains sharp contours and object integrity. This superior performance demonstrates the effectiveness of our proposed MFS and TAA modules in stabilizing motion features and achieving precise spatiotemporal alignment.

Furthermore, Fig.~\ref{fig:vis_seq_comparison} illustrates the temporal consistency and tracking stability of our approach across sampled continuous video sequences. To comprehensively validate our model, we visualize five distinct scenarios. 
The first sequence (\textit{jumping frog}), sampled from the MoCA benchmark, represents typical camouflage and motion. The subsequent four sequences are drawn from our proposed YUV20K dataset, carefully selected to highlight extreme environmental challenges: an underwater \textit{worm} (complex white background), a \textit{nightjar} (intricate texture blending), \textit{birds in snow} (multiple object and color assimilation), and a \textit{safari} scene in the dark (low-light degradation). Note that while each example displays five continuous frames, the temporal strides vary adaptively to capture the specific motion speeds of different targets. 
As visually evident, when handling continuous dynamic changes, baseline methods frequently suffer from severe temporal inconsistency, exhibiting target flickering, boundary collapse. In stark contrast, our framework maintains highly accurate, crisp, and temporally coherent segmentation masks throughout the entire temporal span, powerfully demonstrating the effectiveness of our spatiotemporal alignment and feature stabilization mechanisms.

\begin{table*}[t]
    \centering
    \caption{Ablation study on three datasets. $\downarrow$ indicates lower is better, others are higher is better. The best results are highlighted in \textbf{bold}.}
    \label{tab:ablation_study}
    \setlength{\tabcolsep}{3pt} 
    \resizebox{\textwidth}{!}{
        \begin{tabular}{l | cccc | cccc | cccc}
            \toprule
            \multirow{2}{*}{\textbf{Method}} & 
            \multicolumn{4}{c|}{\textbf{CAD}} & 
            \multicolumn{4}{c|}{\textbf{MoCA-mask}} & 
            \multicolumn{4}{c}{\textbf{YUV20K (Ours)}} \\
            \cmidrule(lr){2-5} \cmidrule(lr){6-9} \cmidrule(lr){10-13}
            
             & $S_m\uparrow$ & $F_m^\omega\uparrow$ & $E_m\uparrow$ & $\mathcal{M}\downarrow$ 
             & $S_m\uparrow$ & $F_m^\omega\uparrow$ & $E_m\uparrow$ & $\mathcal{M}\downarrow$ 
             & $S_m\uparrow$ & $F_m^\omega\uparrow$ & $E_m\uparrow$ & $\mathcal{M}\downarrow$ \\
            \midrule
            
            Baseline 
            & .759 & .562 & .861 & .020 
            & .699 & .421 & .696 & .008 
            & .823 & .691 & .851 & .036 \\
            
            + MFS   
            & .757 & .543 & .852 & .021 
            & \textbf{.715} & .438 & .722 & \textbf{.008} 
            & .807 & .467 & .863 & .032 \\

            + TAA w/ vanilla 3D
            & .762 & .573 & \textbf{.877} & \textbf{.019} 
            & .707 & .443 & .715 & \textbf{.008} 
            & .808 & .664 & .860 & \textbf{.029} \\
            
            + TAA
            & .763 & .575 & .866 & \textbf{.019} 
            & \textbf{.715} & \textbf{.459} & .728 & \textbf{.007} 
            & .805 & .657 & .867 & .030 \\
            \midrule
            \textbf{Ours}
            & \textbf{.771} & \textbf{.584} & .871 & \textbf{.019} 
            & .700 & .432 & \textbf{.733} & \textbf{.008} 
            & \textbf{.858} & \textbf{.780} & \textbf{.920} & .035 \\
            
            \bottomrule
        \end{tabular}
    }
\end{table*}

\subsection{Ablation Study}
In this section, we perform a comprehensive ablation analysis to investigate the contribution of each proposed component. To rigorously evaluate the cross-domain generalization capability of these modules, all experiments in this section are conducted using models trained \textbf{exclusively on the MoCA-Mask training set}. The quantitative results on CAD, MoCA-Mask, and our YUV20K benchmarks are summarized in Tab.~\ref{tab:ablation_study}.

\noindent\textbf{Synergy of MFS and TAA.} 
As shown in Tab.~\ref{tab:ablation_study}, introducing either the Motion Feature Stabilization (MFS) or the Trajectory-Aware Alignment (TAA) module individually brings noticeable improvements to the in-domain MoCA-mask dataset compared to the baseline. However, when deployed to the complex and unseen YUV20K dataset, deploying a single module occasionally leads to performance fluctuations (e.g., a drop in $F_m^\omega$). This suggests that while individual modules excel in MoCA bias motion, single-axis optimization is insufficient to bridge the vast domain gap toward wild, erratic deformations.

Crucially, when both modules are integrated into our \textbf{Full Model} (last row), the cross-domain performance on YUV20K experiences a massive surge ($S_m$ jumps to \textbf{.858} and $F_m^\omega$ to \textbf{.780}). This compelling evidence proves that MFS and TAA are highly complementary. The appearance stabilization provided by MFS acts as a prerequisite for TAA to accurately predict trajectory offsets, while TAA perfectly aligns the stabilized features. Together, they form an indispensable synergy that successfully conquers severe spatiotemporal complex scenarios.

\noindent\textbf{Impact of 3D-CDC in TAA.}
To justify the design choice of using 3D-CDC within the TAA module, we conduct an internal ablation by substituting it with a standard vanilla 3D convolution (denoted as `+ TAA w/o 3D-CDC' in Tab.~\ref{tab:ablation_study}). 

Comparing the 3rd and 4th rows, we observe an intriguing phenomenon. On the source domain (MoCA-Mask), the variant without 3D-CDC performs competitively. However, when generalizing to the YUV20K dataset, removing the 3D-CDC operator leads to a noticeable performance drop. This degradation suggests that standard intensity-based 3D convolutions struggle to distinguish subtle, non-rigid motion patterns from static backgrounds in unseen environments. In contrast, by explicitly leveraging gradient-level differential cues, our full TAA module (equipped with 3D-CDC) captures fine-grained motion dynamics and guides the trajectory alignment significantly more effectively under extreme conditions.

\subsection{Hyperparameter Analysis}
To further investigate the properties of our proposed framework, we conduct detailed analyses on two critical hyperparameters: the trade-off coefficient $\theta$ in the 3D-CDCST module and the dimension of Semantic Basis Primitives (SBP) in the MFS module. 

\noindent\textbf{Impact of $\theta$ in 3D-CDCST.} 
In our 3D-CDCST module, the parameter $\theta$ regulates the contribution ratio between the standard 3D convolution and the central difference convolution. Interestingly, empirical results reveal that the optimal $\theta$ value is highly dependent on the intrinsic motion characteristics of the training domain. Specifically, when optimized on the MoCA dataset, which features relatively simple scenarios, setting $\theta = 0.458$ yields the best performance. However, on our highly complex YUV20K dataset with erratic and severe deformations, the optimal value shifts to $\theta = 0.158$. This discrepancy indicates that different motion distributions require distinct levels of gradient-level differential cues. Motivated by this insight, expanding $\theta$ into a dynamically learnable parameter to further enhance scene-adaptive generalization remains a promising direction for our future work.

\noindent\textbf{Dimension of Semantic Basis Primitives.} 
The dimension of the Semantic Basis Primitives, denoted as $K$, dictates the representational capacity for appearance stabilization in the MFS module. A smaller dimension fails to provide sufficient semantic anchors to comprehensively encompass the complex diversity of camouflaged objects and highly textured backgrounds. Conversely, an excessively large dimension not only introduces burdensome computational overhead but also provokes feature redundancy and over-smoothing, which dilutes critical discriminative cues. Extensive experiments demonstrate that setting the dimension to $K=384$ strikes the perfect balance. This configuration grants the network ample capacity to robustly anchor features against severe motion dynamics without suffering from over-parameterization.

\section{Conclusion}
In this paper, we addressed the critical challenges in Video Camouflaged Object Detection (VCOD) stemming from data scarcity and complex motion scenarios. We introduced YUV20K, a comprehensive and diverse benchmark that fills the gap in representing wild biological behaviors, offering a higher proportion of complex scenarios such as occlusion and hunting behavior compared to prior datasets.
To effectively process these challenging inputs, we proposed a unified spatiotemporal framework designed to rectify motion-induced feature instability and temporal misalignment. Our approach integrates the Motion Feature Stabilization (MFS) module, which leverages Semantic Basis Primitives to anchor semantic features, and the Trajectory-Aware Alignment (TAA) module, which utilizes trajectory-guided deformable sampling for precise feature alignment and aggregation. 
Experimental results confirm that our framework sets a new baseline for the field. Notably, it consistently achieves state-of-the-art performance not only under standard in-domain evaluations but also in rigorous cross-domain settings, exhibiting exceptional generalization capabilities on our challenging YUV20K dataset. We believe this work significantly advances VCOD research from both data and algorithmic perspectives, paving the way for the development of more resilient spatiotemporal reasoning systems in wild, complex scenarios.

\subsection*{Abbreviations}
SBP, semantic basis primitives; MFS, motion feature stabilization; 3D-CDCST, 3 dimension center-difference convolution spatial temporal; TAA, trajectory aware alignment.

\subsection*{Author Contributions}
Y.L. and Y.S. contributed equally to this work. Y.L. conceived the original idea and implemented the main methodology. Y.S. performed partial experiments and contributed to drafting the manuscript. C.H. provided guidance and critically revised the manuscript. Z.Y. supervised the project, organized the research discussions, and finalized the writing. All authors read and approved the final manuscript.

\subsection*{Funding}
This work was supported by the National Natural Science Foundation of China (Grant No. 62576076). The computational resources are supported by SongShan Lake HPC Center (SSL-HPC) in Great Bay University.

\subsection*{Data Availability}
The datasets supporting this study's findings are publicly accessible and include two camouflaged video datasets: CAD \cite{bideau2016scad} (\href{https://drive.google.com/file/d/1XhrC6NSekGOAAM7osLne3p46pj1tLFdI/view?usp=sharing}{Google Drive}), MoCA-Mask \cite{cheng2022implicit} (\href{https://drive.google.com/file/d/1FB24BGVrPOeUpmYbKZJYL5ermqUvBo_6/view?usp=sharing}{Google Drive}). Our YUV20K will be available at \url{https://github.com/K1NSA/YUV20K}.

\section*{Declarations}
\subsection*{Competing interests}
The authors declare no competing interests.

\bibliographystyle{unsrt}
\bibliography{reference}

\begin{thebibliography}{10}

\bibitem{fan2021concealed}
Deng-Ping Fan, Ge-Peng Ji, Ming-Ming Cheng, and Ling Shao.
\newblock Concealed object detection.
\newblock {\em IEEE transactions on pattern analysis and machine intelligence}, 44(10):6024--6042, 2021.

\bibitem{fan2020pranet}
Deng-Ping Fan, Ge-Peng Ji, Tao Zhou, Geng Chen, Huazhu Fu, Jianbing Shen, and Ling Shao.
\newblock Pranet: Parallel reverse attention network for polyp segmentation.
\newblock In {\em International conference on medical image computing and computer-assisted intervention}, pages 263--273. Springer, 2020.

\bibitem{fan2020inf}
Deng-Ping Fan, Tao Zhou, Ge-Peng Ji, Yi~Zhou, Geng Chen, Huazhu Fu, Jianbing Shen, and Ling Shao.
\newblock Inf-net: Automatic covid-19 lung infection segmentation from ct images.
\newblock {\em IEEE transactions on medical imaging}, 39(8):2626--2637, 2020.

\bibitem{rustia2020application}
Dan Jeric~Arcega Rustia, Chien~Erh Lin, Jui-Yung Chung, Yi-Ji Zhuang, Ju-Chun Hsu, and Ta-Te Lin.
\newblock Application of an image and environmental sensor network for automated greenhouse insect pest monitoring.
\newblock {\em Journal of Asia-Pacific Entomology}, 23(1):17--28, 2020.

\bibitem{lamdouar2020betrayed}
Hala Lamdouar, Charig Yang, Weidi Xie, and Andrew Zisserman.
\newblock Betrayed by motion: Camouflaged object discovery via motion segmentation.
\newblock In {\em Proceedings of the Asian conference on computer vision}, 2020.

\bibitem{yang2021self}
Charig Yang, Hala Lamdouar, Erika Lu, Andrew Zisserman, and Weidi Xie.
\newblock Self-supervised video object segmentation by motion grouping.
\newblock In {\em Proceedings of the IEEE/CVF International Conference on Computer Vision}, pages 7177--7188, 2021.

\bibitem{perazzi2016benchmarkmotionblur}
Federico Perazzi, Jordi Pont-Tuset, Brian McWilliams, Luc Van~Gool, Markus Gross, and Alexander Sorkine-Hornung.
\newblock A benchmark dataset and evaluation methodology for video object segmentation.
\newblock In {\em Proceedings of the IEEE conference on computer vision and pattern recognition}, pages 724--732, 2016.

\bibitem{yang2021uncertainty}
Fan Yang, Qiang Zhai, Xin Li, Rui Huang, Ao~Luo, Hong Cheng, and Deng-Ping Fan.
\newblock Uncertainty-guided transformer reasoning for camouflaged object detection.
\newblock In {\em Proceedings of the IEEE/CVF international conference on computer vision}, pages 4146--4155, 2021.

\bibitem{wang2019edvr}
Xintao Wang, Kelvin~CK Chan, Ke~Yu, Chao Dong, and Chen Change~Loy.
\newblock Edvr: Video restoration with enhanced deformable convolutional networks.
\newblock In {\em Proceedings of the IEEE/CVF conference on computer vision and pattern recognition workshops}, pages 0--0, 2019.

\bibitem{dcnzhu2020deformabledetr}
Xizhou Zhu, Weijie Su, Lewei Lu, Bin Li, Xiaogang Wang, and Jifeng Dai.
\newblock Deformable detr: Deformable transformers for end-to-end object detection.
\newblock {\em arXiv preprint arXiv:2010.04159}, 2020.

\bibitem{wang2023internimage}
Wenhai Wang, Jifeng Dai, Zhe Chen, Zhenhang Huang, Zhiqi Li, Xizhou Zhu, Xiaowei Hu, Tong Lu, Lewei Lu, Hongsheng Li, et~al.
\newblock Internimage: Exploring large-scale vision foundation models with deformable convolutions.
\newblock In {\em Proceedings of the IEEE/CVF conference on computer vision and pattern recognition}, pages 14408--14419, 2023.

\bibitem{bideau2016scad}
Pia Bideau and Erik Learned-Miller.
\newblock It’s moving! a probabilistic model for causal motion segmentation in moving camera videos.
\newblock In {\em European Conference on Computer Vision}, pages 433--449. Springer, 2016.

\bibitem{cheng2022implicit}
Xuelian Cheng, Huan Xiong, Deng-Ping Fan, Yiran Zhong, Mehrtash Harandi, Tom Drummond, and Zongyuan Ge.
\newblock Implicit motion handling for video camouflaged object detection.
\newblock In {\em Proceedings of the IEEE/CVF Conference on Computer Vision and Pattern Recognition}, pages 13864--13873, 2022.

\bibitem{xiao2024survey}
Fengyang Xiao, Sujie Hu, Yuqi Shen, Chengyu Fang, Jinfa Huang, Chunming He, Longxiang Tang, Ziyun Yang, and Xiu Li.
\newblock A survey of camouflaged object detection and beyond.
\newblock {\em arXiv preprint arXiv:2408.14562}, 2024.

\bibitem{hall2013camouflage}
Joanna~R Hall, Innes~C Cuthill, Roland Baddeley, Adam~J Shohet, and Nicholas~E Scott-Samuel.
\newblock Camouflage, detection and identification of moving targets.
\newblock {\em Proceedings of the Royal Society B: Biological Sciences}, 280(1758):20130064, 2013.

\bibitem{fan2020camouflaged}
Deng-Ping Fan, Ge-Peng Ji, Guolei Sun, Ming-Ming Cheng, Jianbing Shen, and Ling Shao.
\newblock Camouflaged object detection.
\newblock In {\em Proceedings of the IEEE/CVF conference on computer vision and pattern recognition}, pages 2777--2787, 2020.

\bibitem{cong2023frequency}
Runmin Cong, Mengyao Sun, Sanyi Zhang, Xiaofei Zhou, Wei Zhang, and Yao Zhao.
\newblock Frequency perception network for camouflaged object detection.
\newblock In {\em Proceedings of the 31st ACM international conference on multimedia}, pages 1179--1189, 2023.

\bibitem{sun2024frequency}
Yanguang Sun, Chunyan Xu, Jian Yang, Hanyu Xuan, and Lei Luo.
\newblock Frequency-spatial entanglement learning for camouflaged object detection.
\newblock In {\em European Conference on Computer Vision}, pages 343--360. Springer, 2024.

\bibitem{zhai2021mutual}
Qiang Zhai, Xin Li, Fan Yang, Chenglizhao Chen, Hong Cheng, and Deng-Ping Fan.
\newblock Mutual graph learning for camouflaged object detection.
\newblock In {\em Proceedings of the IEEE/CVF conference on computer vision and pattern recognition}, pages 12997--13007, 2021.

\bibitem{pang2022zoom}
Youwei Pang, Xiaoqi Zhao, Tian-Zhu Xiang, Lihe Zhang, and Huchuan Lu.
\newblock Zoom in and out: A mixed-scale triplet network for camouflaged object detection.
\newblock In {\em Proceedings of the IEEE/CVF Conference on computer vision and pattern recognition}, pages 2160--2170, 2022.

\bibitem{huang2023feature}
Zhou Huang, Hang Dai, Tian-Zhu Xiang, Shuo Wang, Huai-Xin Chen, Jie Qin, and Huan Xiong.
\newblock Feature shrinkage pyramid for camouflaged object detection with transformers.
\newblock In {\em Proceedings of the IEEE/CVF conference on computer vision and pattern recognition}, pages 5557--5566, 2023.

\bibitem{teed2020raft}
Zachary Teed and Jia Deng.
\newblock Raft: Recurrent all-pairs field transforms for optical flow.
\newblock In {\em European conference on computer vision}, pages 402--419. Springer, 2020.

\bibitem{zhang2025explicitemip}
Xin Zhang, Tao Xiao, Ge-Peng Ji, Xuan Wu, Keren Fu, and Qijun Zhao.
\newblock Explicit motion handling and interactive prompting for video camouflaged object detection.
\newblock {\em IEEE Transactions on Image Processing}, 2025.

\bibitem{hui2024endow}
Wenjun Hui, Zhenfeng Zhu, Shuai Zheng, and Yao Zhao.
\newblock Endow sam with keen eyes: Temporal-spatial prompt learning for video camouflaged object detection.
\newblock In {\em Proceedings of the IEEE/CVF conference on computer vision and pattern recognition}, pages 19058--19067, 2024.

\bibitem{pang2024zoomnext}
Youwei Pang, Xiaoqi Zhao, Tian-Zhu Xiang, Lihe Zhang, and Huchuan Lu.
\newblock Zoomnext: A unified collaborative pyramid network for camouflaged object detection.
\newblock {\em IEEE transactions on pattern analysis and machine intelligence}, 46(12):9205--9220, 2024.

\bibitem{meeran2024sampm}
Muhammad~Nawfal Meeran, Bhanu~Pratyush Mantha, et~al.
\newblock Sam-pm: Enhancing video camouflaged object detection using spatio-temporal attention.
\newblock In {\em Proceedings of the IEEE/CVF Conference on Computer Vision and Pattern Recognition}, pages 1857--1866, 2024.

\bibitem{le2019anabranchcamo}
Trung-Nghia Le, Tam~V Nguyen, Zhongliang Nie, Minh-Triet Tran, and Akihiro Sugimoto.
\newblock Anabranch network for camouflaged object segmentation.
\newblock {\em Computer vision and image understanding}, 184:45--56, 2019.

\bibitem{skurowski2018animalchameleon}
Przemys{\l}aw Skurowski, Hassan Abdulameer, Jakub B{\l}aszczyk, Tomasz Depta, Adam Kornacki, and Przemys{\l}aw Kozie{\l}.
\newblock Animal camouflage analysis: Chameleon database.
\newblock {\em Unpublished manuscript}, 2(6):7, 2018.

\bibitem{yunqiu_cod21}
Yunqiu Lyu, Jing Zhang, Yuchao Dai, Aixuan Li, Bowen Liu, Nick Barnes, and Deng-Ping Fan.
\newblock Simultaneously localize, segment and rank the camouflaged objects.
\newblock In {\em Proceedings of the IEEE/CVF Conference on Computer Vision and Pattern Recognition (CVPR)}, 2021.

\bibitem{X-AnyLabeling}
Wei Wang.
\newblock Advanced auto labeling solution with added features.
\newblock \url{https://github.com/CVHub520/X-AnyLabeling}, 2023.

\bibitem{ravi2024sam}
Nikhila Ravi, Valentin Gabeur, Yuan-Ting Hu, Ronghang Hu, Chaitanya Ryali, Tengyu Ma, Haitham Khedr, Roman R{\"a}dle, Chloe Rolland, Laura Gustafson, et~al.
\newblock Sam 2: Segment anything in images and videos.
\newblock {\em arXiv preprint arXiv:2408.00714}, 2024.

\bibitem{ren2024groundedgroundsam}
Tianhe Ren, Shilong Liu, Ailing Zeng, Jing Lin, Kunchang Li, He~Cao, Jiayu Chen, Xinyu Huang, Yukang Chen, Feng Yan, et~al.
\newblock Grounded sam: Assembling open-world models for diverse visual tasks.
\newblock {\em arXiv preprint arXiv:2401.14159}, 2024.

\bibitem{wang2022pvtv2}
Wenhai Wang, Enze Xie, Xiang Li, Deng-Ping Fan, Kaitao Song, Ding Liang, Tong Lu, Ping Luo, and Ling Shao.
\newblock Pvt v2: Improved baselines with pyramid vision transformer.
\newblock {\em Computational visual media}, 8(3):415--424, 2022.

\bibitem{espinosa2022concept}
Mateo Espinosa~Zarlenga, Pietro Barbiero, Gabriele Ciravegna, Giuseppe Marra, Francesco Giannini, Michelangelo Diligenti, Zohreh Shams, Frederic Precioso, Stefano Melacci, Adrian Weller, et~al.
\newblock Concept embedding models: Beyond the accuracy-explainability trade-off.
\newblock {\em Advances in neural information processing systems}, 35:21400--21413, 2022.

\bibitem{koh2020concept}
Pang~Wei Koh, Thao Nguyen, Yew~Siang Tang, Stephen Mussmann, Emma Pierson, Been Kim, and Percy Liang.
\newblock Concept bottleneck models.
\newblock In {\em International conference on machine learning}, pages 5338--5348. PMLR, 2020.

\bibitem{ye_ika2:_2026}
Shuo Ye, Lixin Chen, Qiaoqi Li, Jiayu Zhang, Chaomeng Chen, and Shutao Xia.
\newblock {IKA2}: {Internal} {Knowledge} {Adaptive} {Activation} for {Robust} {Recognition} in {Complex} {Scenarios}.
\newblock {\em Machine Intelligence Research}, 23(2):429--443, April 2026.

\bibitem{yu2020searching3dcdc}
Zitong Yu, Chenxu Zhao, Zezheng Wang, Yunxiao Qin, Zhuo Su, Xiaobai Li, Feng Zhou, and Guoying Zhao.
\newblock Searching central difference convolutional networks for face anti-spoofing.
\newblock In {\em Proceedings of the IEEE/CVF conference on computer vision and pattern recognition}, pages 5295--5305, 2020.

\bibitem{yu2021dual}
Zitong Yu, Yunxiao Qin, Hengshuang Zhao, Xiaobai Li, and Guoying Zhao.
\newblock Dual-cross central difference network for face anti-spoofing.
\newblock {\em arXiv preprint arXiv:2105.01290}, 2021.

\bibitem{pang2020pysodmetrics}
Youwei Pang.
\newblock Pysodmetrics: A simple and efficient implementation of sod metrcis, 2020.

\bibitem{Smeasure}
Deng-Ping Fan, Ming-Ming Cheng, Yun Liu, Tao Li, and Ali Borji.
\newblock Structure-measure: A new way to evaluate foreground maps.
\newblock In {\em ICCV}, pages 4548--4557, 2017.

\bibitem{achanta2009frequencyfmeasure}
Radhakrishna Achanta, Sheila Hemami, Francisco Estrada, and Sabine Susstrunk.
\newblock Frequency-tuned salient region detection.
\newblock In {\em 2009 IEEE conference on computer vision and pattern recognition}, pages 1597--1604. IEEE, 2009.

\bibitem{wFmeasure}
Ran Margolin, Lihi Zelnik-Manor, and Ayellet Tal.
\newblock How to evaluate foreground maps?
\newblock In {\em CVPR}, pages 248--255, 2014.

\bibitem{Emeasure}
Deng-Ping {Fan}, Cheng {Gong}, Yang {Cao}, Bo~{Ren}, Ming-Ming {Cheng}, and Ali {Borji}.
\newblock Enhanced-alignment measure for binary foreground map evaluation.
\newblock In {\em IJCAI}, pages 698--704, 2018.

\bibitem{MAE}
Federico Perazzi, Philipp Kr{\"a}henb{\"u}hl, Yael Pritch, and Alexander Hornung.
\newblock Saliency filters: Contrast based filtering for salient region detection.
\newblock In {\em CVPR}, pages 733--740, 2012.

\bibitem{hui2024implicit}
Wenjun Hui, Zhenfeng Zhu, Guanghua Gu, Meiqin Liu, and Yao Zhao.
\newblock Implicit-explicit motion learning for video camouflaged object detection.
\newblock {\em IEEE Transactions on Multimedia}, 26:7188--7196, 2024.

\end{thebibliography}

\end{document}